\documentclass[lettersize,journal]{IEEEtran}
\usepackage{amsmath,amsfonts,amssymb}

\usepackage{algorithm}  
\usepackage[spaceRequire=true,indLines=false]{algpseudocodex}
\usepackage{algpseudocode}

\usepackage{array}
\newcolumntype{M}[1]{>{\centering\arraybackslash}m{#1}}

\usepackage{textcomp}
\usepackage{stfloats}
\usepackage{url}
\usepackage{verbatim}
\usepackage{graphicx}
\usepackage{subfig}

\def\BibTeX{{\rm B\kern-.05em{\sc i\kern-.025em b}\kern-.08em
    T\kern-.1667em\lower.7ex\hbox{E}\kern-.125emX}}
\usepackage{balance}
\usepackage{tabularx}
\usepackage{cite}
\usepackage{makecell}
\usepackage{multirow}

\makeatletter
\renewcommand{\maketag@@@}[1]{\hbox{\m@th\normalsize\normalfont#1}}%
\makeatother

\begin{document}
\title{CT-VoxelMap: Efficient Continuous-Time LiDAR-Inertial Odometry with Probabilistic Adaptive Voxel Mapping}
\author{Lei Zhao, Xingyi Li, Tianchen Deng, Chuan Cao, Han Zhang, Weidong Chen
	
%\thanks{Manuscript created October, 2020; This work was developed by the IEEE Publication Technology Department. This work is distributed under the \LaTeX \ Project Public License (LPPL) ( http://www.latex-project.org/ ) version 1.3. A copy of the LPPL, version 1.3, is included in the base \LaTeX \ documentation of all distributions of \LaTeX \ released 2003/12/01 or later. The opinions expressed here are entirely that of the author. No warranty is expressed or implied. User assumes all risk.}
}

%\markboth{Journal of \LaTeX\ Class Files,~Vol.~18, No.~9, September~2020}%
%{How to Use the IEEEtran \LaTeX \ Templates}

\maketitle

\begin{abstract}
Maintaining stable and accurate localization during fast motion or on rough terrain remains highly challenging for mobile robots with onboard resources. Currently, multi-sensor fusion methods based on continuous-time representation offer a potential and effective solution to this challenge. Among these, spline-based methods provide an efficient and intuitive approach for continuous-time representation. Previous continuous-time odometry works based on B-splines either treat control points as variables to be estimated or perform estimation in quaternion space, which introduces complexity in deriving analytical Jacobians and often overlooks the fitting error between the spline and the true trajectory over time. To address these issues, we first propose representing the increments of control points on matrix Lie groups as variables to be estimated. Leveraging the feature of the cumulative form of B-splines, we derive a more compact formulation that yields simpler analytical Jacobians without requiring additional boundary condition considerations. Second, we utilize forward propagation information from IMU measurements to estimate fitting errors online and further introduce a hybrid feature-based voxel map management strategy, enhancing system accuracy and robustness. Finally, we propose a re-estimation policy that significantly improves system computational efficiency and robustness. The proposed method is evaluated on multiple challenging public datasets, demonstrating superior performance on most sequences. Detailed ablation studies are conducted to analyze the impact of each module on the overall pose estimation system.

\end{abstract}

\begin{IEEEkeywords}
Continuous-time Odometry, Mapping, Sensor fusion.
\end{IEEEkeywords}

\section{Introduction}

Faced with the increasing application of various types of robots—such as mobile robots, humanoid robots, quadruped robots, wheel-legged robots, and unmanned aerial vehicles—in domains including autonomous driving, embodied intelligence, inspection, and search and rescue, enabling these robots to leverage their onboard resources to maintain stable and reliable motion or pose estimation under conditions of jolts and rapid motion caused by complex terrain or their own complex locomotion patterns is a critical and highly challenging task \cite{onboard-loc, slope-lidar, lidar-nav-sr, rolo-slam, MNE-SLAM}.
LiDAR sensors are widely adopted for pose estimation due to their advantages in measurement range, distance accuracy, and robustness under varying illumination conditions, making them well-suited for complex real-world environments. Furthermore, integrating complementary sensor modalities such as camera, IMU, and wheel odometer \cite{Lidar-imu-cali, liwo, LIC-Fusion, LIC-Fusion2, FAST-LIVO} can further enhance the accuracy and robustness of pose estimation systems. This multi-sensor fusion paradigm has gained significant traction in recent years and has substantially advanced the development of state estimation technologies.

\begin{figure}[htbp]
	\centerline{\includegraphics[width=1.0\columnwidth ]{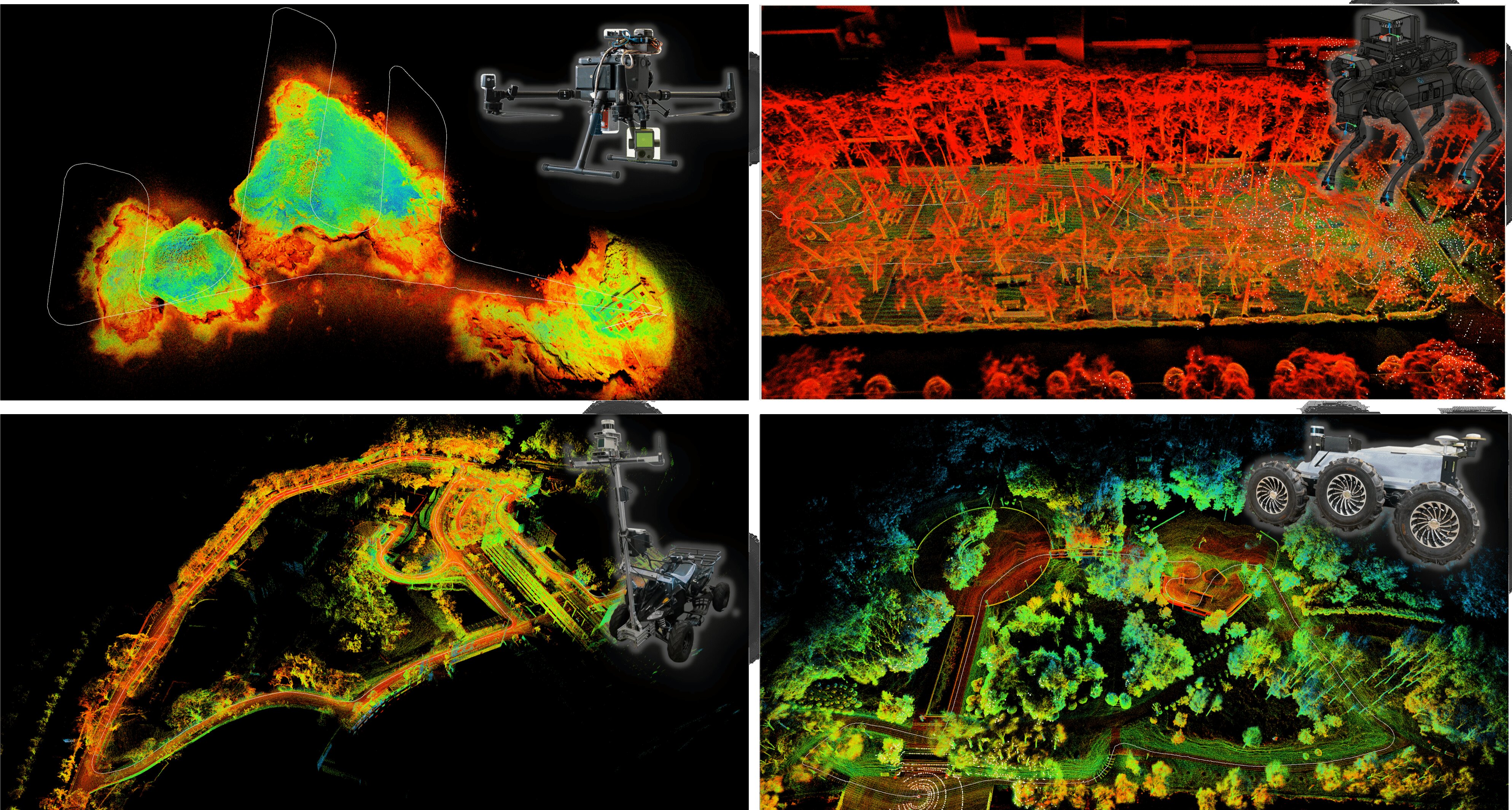}}
	\caption{Dataset platforms and mapping performance. Despite the varying motion patterns and sensor configurations across different dataset platforms, the proposed method achieves robust and accurate pose estimation in all cases.}
	\label{dataset_platform}
\end{figure}

Current popular pose estimation techniques typically operate in discrete time at fixed rates, as exemplified by systems such as FAST-LIO \cite{FAST-LIO}, LeGO-LOAM \cite{lego-loam}, and FAST-LIVO2 \cite{FAST-LIVO2}. These approaches generally require additional preprocessing steps, including sensor time synchronization and LiDAR point cloud distortion (motion compensation), followed by pose estimation within filter or nonlinear optimization frameworks. For filter-based methods, the Iterated Extended Kalman Filter (IEKF) \cite{LIC-Fusion2} and the Iterated Error-State Kalman Filter (IESKF) \cite{FAST-LIO, FAST-LIVO2} are commonly adopted. In such frameworks, the state is first predicted—either under constant velocity assumptions (for LiDAR-only setups) or using IMU measurements (for multi-sensor setups involving IMU)—while simultaneously performing motion compensation to correct LiDAR scan distortions. Maximum a posteriori (MAP) or maximum likelihood estimation (MLE) is then conducted based on the predicted state. For optimization-based frameworks, a sliding window strategy is typically employed, where measurements over a certain time span are used to construct residual terms, and nonlinear optimization is performed under MAP or MLE criteria \cite{CLIC, lio-sam}. In general, optimization-based approaches tend to achieve higher accuracy, while filter-based methods offer greater computational efficiency. However, these discrete-time approaches face limitations when dealing with multi-sensor fusion scenarios, particularly those involving dense, asynchronous sensor measurements, high-fidelity processing requirements, and the need to accommodate varying sensor rates. As pose estimation technology advances, continuous-time trajectory representation-based methods have gradually gained attention as a means to further improve estimation accuracy, robustness, and multi-sensor fusion efficiency.

Continuous-time pose estimation has garnered increasing attention in recent years. Unlike discrete-time approaches, continuous-time methods represent the state to be estimated using continuous functions. A key advantage lies in the ability to query states at arbitrary timestamps via interpolation, which facilitates lossless, dense, and asynchronous multi-sensor fusion. Current continuous-time methods can be broadly categorized into three classes: piecewise linear trajectory-based methods, Gaussian process (GP)-based methods, and B-spline-based methods (particularly cubic B-splines).

Piecewise linear methods typically adopt constant velocity assumptions and represent trajectories using connected linear segments \cite{traj-lo, CT-ICP}. While convenient, their representational capacity for complex motions is limited. In contrast, GP-based and spline-based methods offer greater flexibility in modeling complex motion patterns \cite{continuous-survey}. GP-based approaches model continuous trajectories using Gaussian processes, enabling inference of pose means and covariances at arbitrary timestamps. However, early GP methods required solving high-dimensional, dense inverse matrix for interpolation, incurring high computational costs unsuitable for real-time SLAM \cite{GP-slam}. Subsequent work \cite{sparse-gp-slam} introduced motion priors derived from linear time-varying stochastic differential equations, yielding exactly sparse inverse kernel matrix and improving query efficiency. This was later extended to matrix Lie groups \cite{sparse-gp-matrix-liegroup}.

Compared to GP representation, B-spline-based implementations are more intuitive and flexible for pose estimation tasks. While B-splines can be recursively constructed via the De Boor-Cox formulation \cite{cox1972numerical, de1972calculating}, practical applications predominantly employ cumulative-form B-splines. \cite{cumulative-Bspline} analyzed time derivatives on Lie groups and proposed recursive formulations that improved Jacobian computation efficiency, significantly advancing the application of cubic B-splines in continuous-time pose estimation. Works such as \cite{CLINS, CLIC, SLICT2} represent 6-DoF poses using cumulative cubic B-splines on matrix Lie groups within optimization frameworks. Distinguishing features include loop closure integration in \cite{CLINS} for improved large-scale mapping consistency, camera measurements incorporation in \cite{CLIC} for enhanced accuracy, and analytic Jacobians in \cite{SLICT2} replacing Ceres Solver for improved efficiency.

Despite achieving satisfactory efficiency, these optimization-based frameworks directly treat control points as estimation variables, resulting in relatively complex Jacobian derivations and additional boundary condition considerations, leaving room for further efficiency improvement \cite{RESPLE}. \cite{RESPLE} extended continuous-time B-spline estimation to filter frameworks for enhanced efficiency, albeit using quaternion representations that complicate Jacobian derivation and lack intuitiveness. Furthermore, none of the aforementioned B-spline methods account for fitting errors between splines and true trajectories. In summary, while B-spline methods offer flexibility and intuitiveness, they lack uncertainty characterization and warrant further theoretical development.

Motivation and Contribution: To further enhance the efficiency and performance of B-spline-based continuous-time pose estimation, the main contributions of this paper are summarized as follows:

\begin{itemize}
	\item We propose an IEKF-based multi-sensor fusion odometry framework that integrates a more consistent spline representation with a hybrid-feature voxel map, achieving both high efficiency and robustness. The proposed method attains superior performance on multiple challenging public datasets while maintaining computational efficiency and system robustness.
	\item Building upon \cite{cumulative-Bspline}, we derive a more compact cumulative-form cubic B-spline representation on Lie groups. By treating control point increments—rather than the control points themselves—as estimation variables, we obtain simplified analytic Jacobians that are unified in form and eliminate the need for explicit boundary condition handling.
	\item To account for spline fitting errors, we leverage IMU measurements to estimate such errors online, thereby enhancing system robustness. Furthermore, we adopt a voxel map structure with hybrid features (plane features and voxel features) to improve estimation accuracy.
	\item We introduce a re-estimation policy that decomposes an $m$-dimensional problem into $k$ subproblems of dimension $m/k$, significantly improving both computational efficiency and system robustness.
	\item Extensive experiments on public datasets, along with ablation studies, provide a comprehensive evaluation of the proposed algorithm. The complete code will be open-sourced to contribute to the community.
\end{itemize}

The mapping performance on different public datasets and platforms is shown in Figure \ref{dataset_platform}, where the images of the robot are respectively sourced from the data collection platforms corresponding to the public datasets \cite{MARS_LVIG, M2UD, MCD, Diter++}.

\section{related work}
This section briefly reviews prevalent LiDAR-based and LiDAR-inertial odometry frameworks, with a focus on the evolution from discrete-time to continuous-time pose estimation paradigms.

\textbf{Discrete-time LiDAR-Based Odometry}: early LiDAR-based pose estimation predominantly relied on 2D LiDAR sensors \cite{ldiar-slam-2011}. With the advent of 3D LiDAR technology, methods leveraging 3D measurements have gained increasing attention. LOAM \cite{loam} represents a seminal work in 3D LiDAR SLAM, which extracts geometric features from point clouds and employs nonlinear optimization to achieve favorable accuracy and efficiency. Subsequent works such as \cite{lego-loam} improved feature extraction by incorporating ground features and introducing loop closure, thereby mitigating z-axis drift and enhancing mapping consistency. \cite{EKF-LOAM} further extended the LOAM framework to multi-sensor configurations, improving robustness in degraded environments.

Compared to filtering-based approaches, LOAM-style methods exhibit relatively lower computational efficiency \cite{LIC-Fusion}. In contrast, \cite{LIC-Fusion, LIC-Fusion2}, building upon the MSCKF paradigm \cite{msckf, msckf-2.0}, leverage state augmentation and QR decomposition-based state compression to achieve accuracy comparable to optimization-based methods while maintaining high efficiency. To exploit finer geometric details by utilizing raw point clouds directly \cite{FAST-LIO, FAST-LIO2}, the IESKF framework combined with point-to-plane ICP has demonstrated remarkable efficiency and accuracy. More recent advances \cite{voxel_map, FAST-LIVO2} adopt voxel-based representations and leverage plane features within local maps to further improve both precision and computational performance.

However, with the rapid progress in autonomous driving and embodied intelligence, sensor suites have become increasingly diverse. The limitations of discrete-time pose estimation methods in accommodating asynchronous fusion of multiple homogeneous or heterogeneous sensors have become increasingly evident. Consequently, continuous-time pose estimation techniques have garnered growing research interest.

\textbf{Continuous-time LiDAR-based odometry}: compared to discrete-time methods that estimate poses only at discrete timestamps, continuous-time approaches treat the pose as lying on a continuous trajectory, with poses at specific moments obtained via sampling along this trajectory. Consequently, continuous-time SLAM can also be referred to as Simultaneous Trajectory Estimation and Mapping (STEAM) \cite{sparse-gp-slam}. When measurements acquired at different timestamps arrive, the corresponding states can be queried from the continuous trajectory to formulate residuals, eliminating the need for additional temporal synchronization or motion compensation \cite{CLINS}.

Continuous-time trajectory representations can be broadly categorized into parametric methods (e.g., linear interpolation, splines) and non-parametric methods (e.g., one-dimensional Gaussian processes) \cite{continuous-survey}. Early parametric continuous-time approaches commonly employed linear interpolation or piecewise continuous functions to represent trajectories. For instance, \cite{CT-ICP} utilized spherical linear interpolation for motion compensation during rapid motion, while \cite{traj-lo} adopted piecewise continuous linear functions to represent robot trajectories. Although these methods achieve satisfactory pose estimation performance under relatively fast motions, linear interpolation and piecewise linear functions have limited representational capacity for complex motion patterns and cannot guarantee velocity or acceleration continuity at junction points.

Gaussian process-based and B-spline-based methods address these limitations. \cite{GP-slam} implemented a Gaussian process-based SLAM approach, but querying poses between discrete timestamps required solving high-dimensional, dense matrix inverses, rendering it unsuitable for real-time applications. Subsequent work introduced motion priors derived from linear time-varying stochastic differential equations, yielding sparse inverse kernel matrix and enabling real-time deployment \cite{sparse-gp-slam}. \cite{cte-mlo} further extended this framework to filter paradigms, achieving more efficient implementations. However, Gaussian process-based methods rely heavily on the form of motion priors.

In contrast, B-spline-based methods offer greater intuitiveness and convenience. \cite{CLINS} implemented continuous-time LiDAR-inertial odometry using cumulative-form B-splines derived from \cite{cumulative-Bspline}, albeit with relatively limited computational efficiency. \cite{SLICT2} improved efficiency by introducing analytic Jacobians with respect to control points, replacing Ceres Solver's automatic differentiation, while maintaining high pose estimation accuracy. Nevertheless, further improvements remain possible. \cite{RESPLE} subsequently extended continuous-time estimation to filter frameworks, employing quaternion representations for rotational trajectories and achieving efficient continuous-time estimation, albeit at the cost of increased complexity. Overall, continuous-time pose estimation methods based on B-splines still require further development.

\section{Preliminlaries}
This section presents the framework of the proposed approach, including coordinate system definitions, sensor measurement models, and the cumulative-form continuous-time trajectory representation.
\subsection{System Overview, Coordinate Frames and Mesurement Model}

\begin{figure}[htbp]
	\centerline{\includegraphics[width=1.0\columnwidth ]{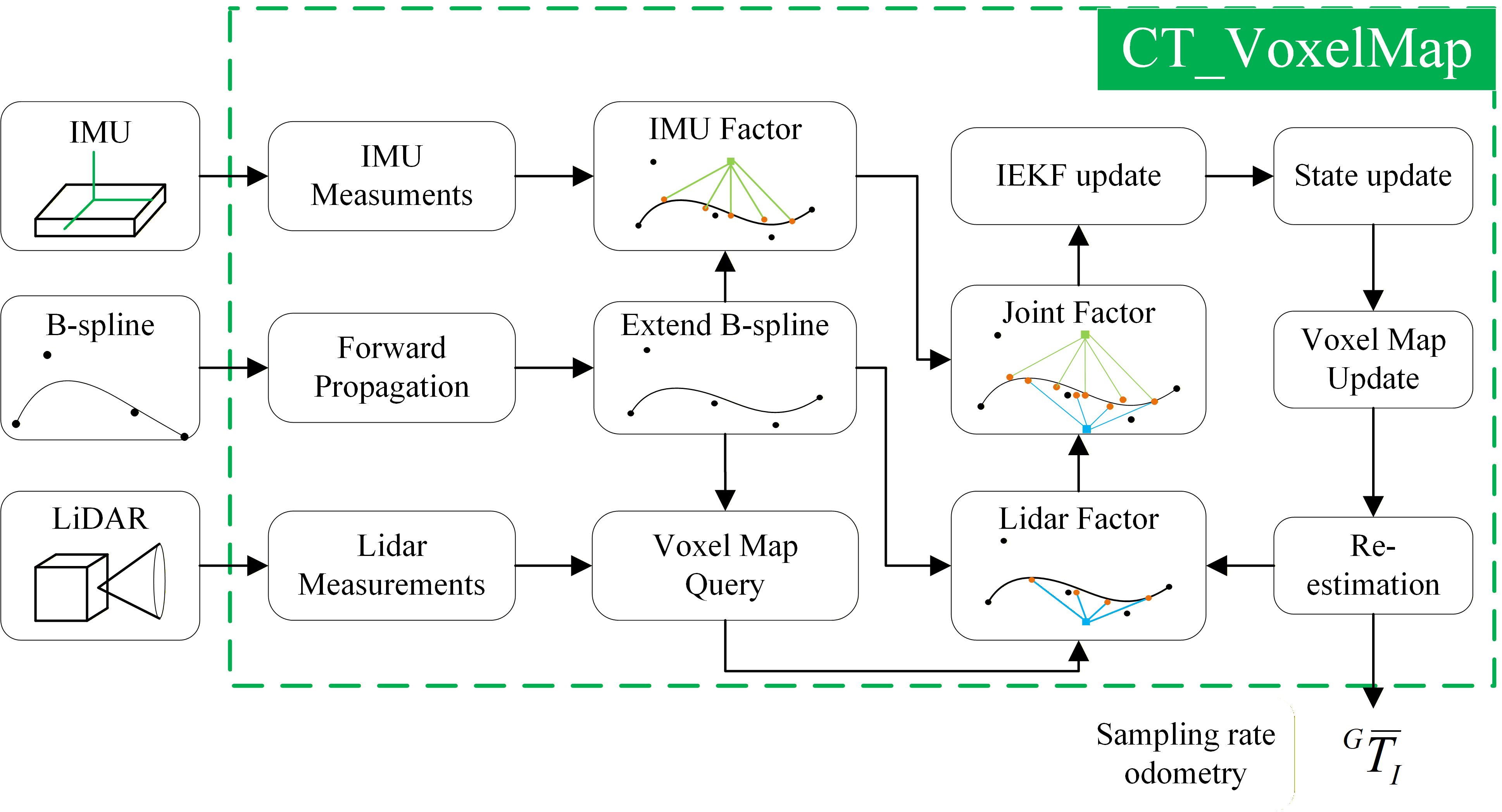}}
	\caption{Schematic diagram of system overview.}
	\label{system_overview}
\end{figure}

The framework of the proposed system is illustrated in Fig.\ref{system_overview}. The LiDAR coordinate frame is denoted as $^{L}(\cdot)$, the IMU coordinate frame as $^{I}(\cdot)$, and the global coordinate frame as $^{G}(\cdot)$. The entire system incorporates two sensors: LiDAR, IMU. The corresponding measurement models are described as follows.

\subsubsection{\textbf{IMU measurement model}}
\begin{align*}
	^{I}\omega_m = {^{I}\omega} + b_{\omega}+n_w, ^{I}a_m = {^{I}a} + b_{a} + {_{G}^{I}R} {^{G}g} + n_{a}.
\end{align*}
where $b_a$ denotes the accelerometer bias, $b_\omega$ denotes the gyroscope bias, $n_w$ and $n_a$ represent the additive white noise in angular velocity and acceleration measurements, respectively. 
%These noise terms are modeled as zero-mean Gaussian white noise: $n_a \sim \mathcal{N}(0,\sigma^{2}_a)$, $n_w \sim \mathcal{N}(0,\sigma^{2}_\omega)$.

\subsubsection{\textbf{LiDAR measurement model}} 
let $\{^{L}P_{j}, j = 1, ..., m \}$ denote the points in the current scan, which are sampled in the local LiDAR coordinate frame $\{L\}$. Due to LiDAR measurement noise, each measured point $^{L}P_{j}$ is typically corrupted by noise $^{L}n_{j}$, which consists of both ranging and beam-directing noise components\cite{FAST-LIO}.

%Removing this noise yields the true point position $^{L}P_{j}^{gt}$ in the LiDAR coordinate frame.
\begin{align*}
	^{L}P_{j}^{gt} = {^{L}P_{j}} + {^{L}n_{j}}
\end{align*}

\subsection{B-spline Based Interpolation}

In recent years, continuous-time odometry methods have achieved remarkable performance \cite{SLICT2, CTA-LO, CLINS, CT-ICP}. Due to the relative simplicity of spline representations, we adopt the cumulative-form B-spline to represent continuous trajectories \cite{continuous-survey}. This representation is characterized by the following parameters: knot interval $\Delta t$, spline order $N$ (or degree $D \triangleq N - 1$), and a set of control points $\{ T_{m}\}, m = 0,...,M$, where $T_{m} \in SE(3)$. Each control point $T_m$ is associated with a knot $t_m$, satisfying $t_m = t_0 + m \Delta t$. We employ cubic splines ($N=4$) with a constant knot interval $\Delta t = \tau$. Unless otherwise specified, all subsequent references to spline trajectories refer to uniform cubic splines. The cumulative form of the cubic B-spline can be described as follows \cite{cumulative-Bspline}:
\begin{equation}
	\begin{aligned}
		&R(t) = R_{i} \prod_{j = 1}^{N-1} Exp( \tilde{\lambda}_{j}(u) Log(R_{i+j-1}^{-1} R_{i+j})), \\
		& p(t) = p_{i} + \sum_{j = 1}^{N-1} \tilde{\lambda}_{j}(u) (p_{i+j} - p_{i+j -1}).
		\label{spline_traj}
	\end{aligned}
\end{equation}

Where $u = \frac{t - t_{i}}{\Delta t}, t \in [ t_i, t_{i+1} ), t_{i} \in \{t_m | m = 0,...,M\}$. The calculation of $\tilde{\lambda}_{i}$ and its first and second derivatives is as follows:
\begin{equation}
	\begin{aligned}
	&[\tilde{\lambda}_{0}, \tilde{\lambda}_{1},...,\tilde{\lambda}_{N-1}]^{T} = \tilde{B}^{(N)}[1, u ,...,u^{N-1}]^{T}, \\
	&[\dot{\tilde{\lambda}}_{0}, \dot{\tilde{\lambda}}_{1},...,\dot{\tilde{\lambda}}_{N-1}]^{T} = \tilde{B}^{(N)}[0, 1 ,...,\dot{u}^{N-2}]^{T}, \\
	&[\ddot{\tilde{\lambda}}_{0}, \ddot{\tilde{\lambda}}_{1},...,\ddot{\tilde{\lambda}}_{N-1}]^{T} = \tilde{B}^{(N)}[0, 0 ,...,\ddot{u}^{N-3}]^{T}.
	\label{Bspline_lambda}
	\end{aligned}
\end{equation}

Where $\dot{u}^{j} = \frac{j}{\Delta t} u^{j-1}$ and $\ddot{u}^{j} = \frac{j(j-1)}{\Delta t^2} u^{j-2}$. Given $N$, a blending matrix $B^{(N)}$ and its cumulative form $\tilde{B}^{(N)}$ can be found in \cite{SLICT2, cumulative-Bspline}.

% \begin{align*}
% 	&B^{(N)} = [b_{s,k}] \in R^{N \times N}; s,k \in {0,1,...,N-1}, \\
% 	&\tilde{B}^{(N)} = [\tilde{b}_{s,k}] = \frac{1}{6}\begin{bmatrix}
% 		6 & 0 & 0 & 0 \\
% 		5 & 3 & -3 & 1 \\
% 		1 & 3 & 3 & -2 \\
% 		0 & 0 & 0 & 1
% 	\end{bmatrix}, \tilde{b}_{s,k} = \sum_{j = s}^{D} b_{j,k}, \\
% 	&b_{s,k} = \frac{1}{k!(D-k)!} \sum_{l = s}^{D}(-1)^{l-s} C^{l-s}_{N}(D-l)^{D-k}. 	
% \end{align*}
%where $ C^{l-s}_{N} = \frac{N!}{(l-s)!(N-l+s)!} $. 

\section{State Estimation}

In the state estimation task employing continuous-time trajectory representation, the state vector $\chi$ comprises the IMU accelerometer bias $b_a$, gyroscope bias $b_w$, gravity $g$, and the control point increments of the B-spline corresponding to the recent continuous trajectory segment.
To facilitate estimation, we examine the rotational control points. From eq.(\ref{spline_traj}) and eq.(\ref{Bspline_lambda}), it can be observed that $\tilde{\lambda}_0(u) = 1$. Consequently, the continuous trajectory representation can be reformulated as:
\begin{small}
\begin{equation}
	\begin{aligned}
		R(t) = R_{i-1} \prod_{j = 0}^{N-1} Exp( \tilde{\lambda}_{j}(u) d_j), d_j = Log(R^{-1}_{i+j-1} R_{i+j})
		\label{rotation}	
	\end{aligned}
\end{equation}
\end{small}

Similar to the approach in \cite{RESPLE}, we estimate $d_j$ rather than directly estimating ${^{G}_{C}R^{T}}$. The key distinction lies in our estimation being performed on the matrix Lie group $SO(3)$, which offers a more intuitive formulation. Based on the above representation, the corresponding recursive formulation for angular velocity can be derived as:
\begin{equation}
	\begin{aligned}
		\omega^{(j+1)} &= A_{j}^T \omega^{(j)} + \dot{\tilde{\lambda}}_{j}(u) d_{j}, j = 0,...,N-1. \\
		\omega(t) &= \omega^{(N)}, \omega^{(0)} = 0, \\
		A_{j} &= Exp(\tilde{\lambda}_{j}(u) d_j), j = 0,...,N-1.
		\label{omega_t}
	\end{aligned}
\end{equation}

The same idea is applied to the positional control points, yielding:
\begin{equation}
	p(t) = p_{i-1} + \sum_{j = 0}^{N - 1} \tilde{\lambda}_{j}(u) d_{j}, d_{j} = p_{i+j} - p_{i+j-1}.
	\label{position}
\end{equation}

The corresponding velocity and acceleration can be computed as follows:
\begin{equation}
	\begin{aligned}
		v(t) = \sum_{j = 0}^{N - 1}  \dot{\tilde{\lambda}}_{j}(u) d_{j}, d_{j} = p_{i+j} - p_{i+j-1}, \\
		a(t) = \sum_{j = 0}^{N - 1}  \ddot{\tilde{\lambda}}_{j}(u) d_{j}, d_{j} = p_{i+j} - p_{i+j-1}.
	\end{aligned}
	\label{velocity_acc}
\end{equation}

The advantages of this formulation are twofold. First, the state variables corresponding to the control points remain in Euclidean space, and the increments are small quantities, ensuring that the state variables stay within a relatively small range, which benefits state estimation. Second, based on the proposed continuous trajectory representation, the boundary conditions are consistent when deriving Jacobian matrices, eliminating the need to separately consider the cases $j=0$ and $j=N-1$, thus resulting in a more compact formulation. Consequently, the corresponding system can be expressed as:
\begin{align*}
	\chi & \doteq  \begin{bmatrix}
		\chi_{C_r} & \chi_{C_p} & \chi_{I} 
	\end{bmatrix}^T \\
	\chi_{C_r} &= [d_{R,j}^T], \chi_{C_p} = [d_{p,j}^T], j = 0,...,N-1, \\
	\chi_I &= [ b_w^T, b_a^T, {^{G}g^T}], \\
	u &\doteq [\omega_m^T, a_m^T]^T, w \doteq [n_{b_w}^T, n_{ba}^T]^T .
\end{align*}

$d_{R,j}^T, d_{p,j}^T, j = 0,...,N-1$ denote the rotational and positional control point increments corresponding to the newly added B-spline segment with their specific definitions given in eq.(\ref{rotation}) and eq.(\ref{position}), respectively. $b_w$, $b_a$, and ${^{G}g}$ represent the gyroscope bias, accelerometer bias, and gravity vector, respectively. $u$ denotes the IMU measurement input, and $w$ represents the corresponding noise.

\subsection{Kinematic Model}

Based on the characteristics of uniform B-splines, the $k$-th spline segment represents the continuous trajectory over the interval $[t_{k,s}, t_{k,e}]$. When a new LiDAR scan arrives, the maximum timestamp of its points is denoted as $t_{L_e}$. If $t_{L_e} \in [t_{k,s}, t_{k,e}]$, the control points remain unchanged during the state prediction stage. However, when $t_{L_e} > t_{k,e}$, new segments and corresponding control points need to be extended to represent the additional trajectory information. Resulting in a jump in the state $\chi$, while $b_w$, $b_a$, and ${^{G}g}$ remain unchanged.

\textbf{hybrid system}: Based on the above description, the state estimation task based on continuous-time representation can be modeled as a hybrid system, where the state undergoes jumps at discrete instants $t_{k}, k = 1,...,n$, while evolving continuously over the intervals $t_i \in \left [ t_{k}, t_{k+1} \right ) , k = 1,...,n$. The system can then be expressed as:
\begin{equation}
	\begin{aligned}
		\left\{\begin{matrix}
			\dot{\chi}(t) = 0 \boxplus w, t \in \left [ t_{k}, t_{k+1} \right ) , k = 1,...,n \\
			\chi_{t^{+}} = A \chi_{t^{-}} , t = t_k, k = 1,...,n.
		\end{matrix}\right.
		\label{hybrid_syetem}
	\end{aligned}
\end{equation}

Where $w$ denotes the process noise, all modeled as Gaussian white noise. (The definitions of $\boxminus$ and $\boxplus$ follow those in \cite{FAST-LIO2}). $A$ represents the state transition matrix, defined as $A = diag(A_r, A_p, I_3, I_3, I_3)$. Here, $A_r$ denotes the state transition matrix corresponding to the rotational control point increments, and $A_p$ denotes the state transition matrix corresponding to the positional control point increments. When no new control points are added, both $A_r$ and $A_p$ are identity matrix. When new control points are added, the specific form needs to be discussed.

For the prediction of control points, due to the local support property of B-splines, the control points corresponding to the newly extended $k+1$-th segment can be initialized from the control points of the $k$-th segment. The corresponding state can be initialized using the state at time $t_{k}^{-}$, yielding:
\begin{align*}
	& b_w^{t_{k}^{+}} = b_w^{t_{k}^{-}}, b_a^{t_{k}^{+}} = b_a^{t_{k}^{-}},  {^{G}g}^{t_{k}^{+}} = {^{G}g}^{t_{k}^{-}},\\
	& d_{R,j}^{t_{k}^{+}} = d_{R,j+1}^{t_{k}^{-}}, d_{p,j}^{t_{k}^{+}} = d_{p,j+1}^{t_{k}^{-}}, j = 0,...,N-2. 
\end{align*}

For the initialization of $d_{R,N-1}^{t_{k}^{+}}$ and $d_{p, N-1}^{t_{k}^{+}}$, it is related to the initialization strategy. The initialization strategy adopted in this paper is the constant velocity initialization strategy, which maintains the linear velocity and angular velocity at times $t_{k}$ and $t_{k+1}$ unchanged. For the rotational control points, according to eq.(\ref{omega_t}), we can obtain:
\begin{align*}
	&\omega(t_{k}) = \omega(t_{k+1}) \\
	&\Rightarrow  A_{N-1}^T\dots A_{0}^T \omega^{(0)} + \dots + \dot{\tilde{\lambda}}_{N-1}(0)d_{N-1} \\
	& = A_{N-1}^T\dots A_{0}^T \omega^{(0)} + \dots + \dot{\tilde{\lambda}}_{N-1}(1)d_{N-1}\\
	&\Rightarrow A_{N-1}^T|_{u = 0} d_{N-3} + d_{N-2} = A_{N-1}^T |_{u = 1} d_{N-2} + d_{N-1} \\
	&\Rightarrow  d_{R, N-1}^{t_{k}^{+}} \simeq d_{R, N-2}^{t_{k}^{-}},
\end{align*}
when the control point frequency is sufficiently high, $d_{j}, j = 0,...,N-1$ are small quantities. For the positional control points, according to eq.(\ref{velocity_acc}), we can obtain:
\begin{align*}
	&v(t_{k+1})= v(t_{k}) \Rightarrow d_{N-3} = d_{N-1} \Rightarrow d_{p, N-1}^{t_{k}^{+}} = d_{p, N-2}^{t_{k}^{-}}.
\end{align*}

Based on the discussion above, when new control points are added, the definitions of $A_r$ and $A_p$ are as follows:
\begin{align*}
	A_r = \begin{bmatrix}
		0 & I_3 & 0 & 0 \\
		0 & 0 & I_3 & 0 \\
		0 & 0 & 0 & I_3 \\
		0 & 0 & I_3 & 0
	\end{bmatrix},  	
	A_p = \begin{bmatrix}
		0 & I_3 & 0 & 0 \\
		0 & 0 & I_3 & 0 \\
		0 & 0 & 0 & I_3 \\
		0 & 0 & I_3 & 0
	\end{bmatrix}
\end{align*}

%\subsubsection{prediction and covariance propagation}
%
%Discretizing the kinematic equations of the continuous system using the forward Euler method according to eq.(\ref{hybrid_syetem}) and unifying them with the jump system yields (where, due to discretization, the system performs estimation every $\Delta t$, thus jumps also occur at intervals of at least $\Delta t$):
%\begin{align*}
%	\chi_{k+1} = A_k \chi_{k} \boxplus w_k.
%\end{align*}
%
%where $w_k = w \Delta t$ is the additive process noise at time $k$, modeled as Gaussian white noise with corresponding covariance $Q_k$, and $A_k$ is the state transition matrix at time $k$. For the discrete continuous-state equation, $A_k = I_{33}$; for the jump system, $A_k = diag(A_r, A_p, I_3, I_3, I_3)$.
%The corresponding state propagation equation is:
%\begin{align}
%	\hat{\chi}_{k+1|k} = A_k \hat{\chi}_{k|k}, \hat{\chi}_{k|k} = \bar{\chi}_{k|k}.
%	\label{state_propagate}
%\end{align}
%
%where $\bar{\chi}_{k|k}$ is the optimal estimate at time $k$. The corresponding covariance propagation is:
%\begin{align}
%	\hat{P}_{k+1 | k} = \tilde{A}_k \hat{P}_{k | k} \tilde{A}_k^T + {Q}_k, \hat{P}_{k | k} = \bar{P}_{k|k}.
%\end{align}
%
%where $\tilde{A}_k = diag(A_r, A_p, I_3, I_3, I_2)$. Since the gravity is considered with its perturbation lying in the tangent space of $\mathbb{S}^2$, the actual perturbation is two-dimensional.
%$\bar{P}_{k | k}$ denotes the covariance at the previous time instant, and $\tilde{Q}_k$ represents the process noise at time $k$.

\subsection{IMU Observation}
Upon the arrival of the $(k+1)$-th LiDAR scan, based on the inter-scan LiDAR point cloud information over $[t_{s,k+1}, t_{e, k+1}]$ and the IMU measurements, maximum a posteriori (MAP) estimation of the state is performed within the iterated extended Kalman filter (IEKF) framework. The corresponding observation equation is formulated as follows:
\begin{align*}
	z_{k} = h(f(\chi_k; t_k)) + v_k.
\end{align*}

Where $v_k$ denotes the observation noise, $h$ represents the corresponding observation equation, and $f$ is a function of $\chi$ that yields the interpolated information obtained from control points at time $t_k \in [t_{k,s}, t_{k,e}]$, such as $R_{t_k}, p_{t_k}$, etc.

%In the iterative update, the current state can be expressed as $\hat{\chi}_{i+1} = \hat{\chi}_i \boxplus \delta \chi_i$, where $\delta \chi$ is the state increment and $\hat{\chi}_{i}$ denotes the estimate obtained at the $i$-th iteration. For variables in $\mathbb{R}^3$, $p_{i+1} = p_i + \delta p_i$, with $p \in \mathbb{R}^3$. For variables in $\mathbb{S}^2$, the update can be performed as follows:
%\begin{align*}
%	s_{i+1} = s_{i} \boxplus \delta = Exp(B_s \delta) s_i, s_i\in S^{2}, \delta \in R^{2}.
%\end{align*}
%
%Where $B_s = [b_1, b_2]_{3 \times 2}$ denote the basis vector on the tangent space of $S^2$ at $s$. The basis $B_s$ is not specified, which can be made arbitrary as long as it forms an orthonormal basis in the tangent plane of $s$. 

\subsubsection{\textbf{IMU observation with continuous-time trajectory}}

\begin{figure}[htbp]
	\centerline{\includegraphics[width=1.0\columnwidth ]{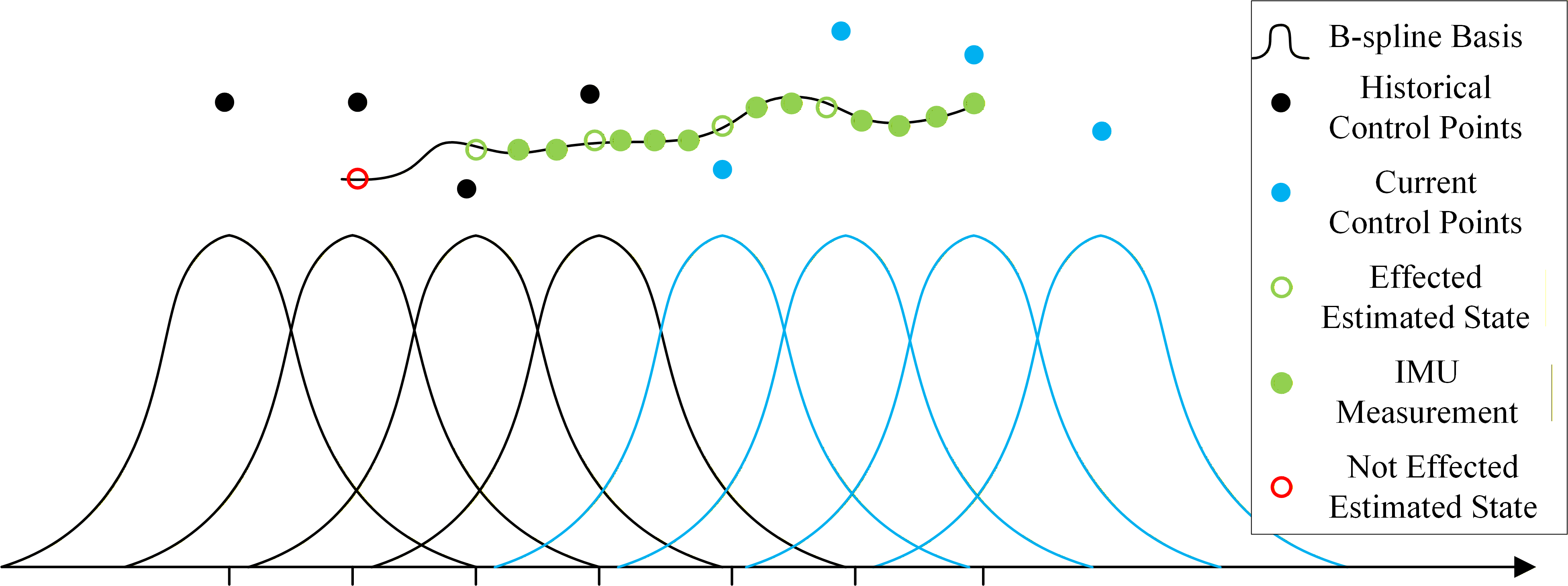}}
	\caption{The black line denotes a continuous trajectory, represented by a 4th-order spline. The solid green circle denotes the IMU measurement, while the hollow green and red circles represent the affected and unaffected historical estimated states, respectively.}
	\label{bspline_trajectory}
\end{figure}

Based on the state information from the prediction stage, to achieve more accurate estimation of the control points, we utilize the IMU measurements, biases, and gravity information over the interval $[t_{s,k+1}, t_{e,k+1}]$.
Furthermore, due to the local support property of B-splines, the control points of the current trajectory segment also influence the representation of historical trajectory segments. Specifically, the $N$ control points of the current segment affect $N$ spline trajectory segments, involving $2N-1$ control points in total. When formulating observations, both measurement and estimation information within the affected intervals are comprehensively considered, i.e., the IMU measurements and estimated biases and gravity information over $[t_{s,k-N+2}, t_{e,k}]$. Each interval has the same $\Delta t$, and $t_{e, i} = t_{s, i+1}, i = 0,...,n$. A schematic illustration is provided in Fig.~\ref{bspline_trajectory}.

Based on the above description, the observation equation formulated from IMU poses and measurements over the interval $[t_{s, k-N+2}, t_{e, k+1}]$ is expressed as follows:
\begin{equation}
\begin{aligned}
	h_B(\chi, 0) &= \begin{bmatrix}
		{^{G}_{I_k}R(t_k)^T}(a(t_k) + {^G\hat{g})} + \hat{b}_{a} - {^{I_{t_k}}a_{m}} \\
		\omega(t_k) + \hat{b}_{\omega} - {^{I_{t_k}}\omega_{m}}\\ 
		\hat{b}_{\omega} - b_{\omega_{t_k}} \\
		\hat{b}_{a} - b_{a_{t_k}} \\
		{^{G}\hat{g}} \boxminus {^{G}g_{t_k}}
	\end{bmatrix}, \\
	&= \begin{bmatrix}
		h_a^T & h_{\omega}^T & h_I^T
	\end{bmatrix}^T .
	\label{Bspline_observation}
\end{aligned}
\end{equation}

Where $R(t)$ and $p(t)$ are computed from eq.(\ref{rotation}) and eq.(\ref{position}), respectively. $\omega(t)$ can be obtained from eq.(\ref{omega_t}), and $a(t)$ can be obtained from eq.(\ref{velocity_acc}).
%The variables $u$ and $t$ are transformed according to eq.(\ref{spline_traj}).
${^{G}g_{t_k}}$ denotes the gravity estimate at time $t_k$, $b_{\omega_{t_k}}$ and $b_{a_{t_k}}$ represent the angular velocity and acceleration biases at time $t_k$, and ${^{I_{t_k}}a_m}$ and ${^{I_{t_k}}\omega_m}$ are the IMU measurements at time $t_k$. ${^{G}\hat{g}}$, $\hat{b}_{a}$, and $\hat{b}_{\omega}$ are the state variables at the current time. The term ${^{G}\hat{g}} \boxminus {^{G}g_{t_k}}$ can be computed on $\mathbb{S}^2$ and can be computed as follows:
\begin{align*}
	s \boxminus d = B_d^T \frac{\left \lfloor d \right \rfloor s}{||\left \lfloor d \right \rfloor s||} atan2(||\left \lfloor d \right \rfloor s||, d^T s), d \neq -s, s, d \in \mathbb{S}^2.
\end{align*}

\textbf{Jacobian}: linearizing the observation equation around $\hat{\chi}_{k+1|k}$ yields:
\begin{align*}
	h_{B}(\chi, n) &= h_B(\hat{\chi} \boxplus \delta \chi, n) \\
	&\simeq h_{B}(\hat{\chi}, 0 ) + H_B \delta \chi + n, n \sim N(0, {\textstyle \sum_{B}})
\end{align*}

Where $\sum_{B}$ is the covariance of the IMU observation noise. Based on the IMU measurement model, the noise in the observation equations $h_a$ and $h_w$ is approximated by the measurement noise. $H$ is the Jacobian matrix obtained when linearizing $h_{B}(\chi, 0)$ around $\hat{\chi}_{k+1|k}$:
\begin{align*}
	H_{B} 
	&= \lim_{\delta \chi \to 0}  \frac{ h_{B}(\hat{\chi} \boxplus \delta \chi, 0) \boxminus h_{B}(\hat{\chi}, 0)}{ \delta \chi} \\
	&= \begin{bmatrix}
		\frac{\partial h_B }{\partial \hat{d}_{R,j}} & \frac{\partial h_B}{ \partial \hat{d}_{p,j} } & \frac{\partial h_B}{\partial \hat{\chi}_I}
	\end{bmatrix}
\end{align*}

Based on the observation equation $h_B(\chi, n)$ (eq.(\ref{Bspline_observation})), the Jacobian matrix is computed using the chain rule. For notational convenience and without ambiguity, we omit the subscript $k$ in the following derivations:
\begin{align*}
	\frac{\partial h_a}{\partial \hat{d}_{R, j} } = \frac{\partial h_a}{\partial R(t)} \frac{\partial R(t)}{ \partial \hat{d}_{R, j}}, j = 0,...,N-1.
\end{align*}

Where $\frac{\partial h_a}{\partial R(t)}$ can be computed as follows:
\begin{align*}
	\frac{\partial h_a}{\partial R(t)} = \left \lfloor {^{G}_{I}R(t)^{T}}(a(t) + {^{G}g})  \right \rfloor_{\times}
\end{align*}

According to eq.(\ref{rotation}), $\frac{\partial R(t)}{ \partial \hat{d}_{R, j}}$ can be computed as follows:
\begin{equation}
\begin{aligned}
	&\frac{\partial R(t)}{ \partial \hat{d}_{R, j}} = \frac{\partial R(t)}{\partial d_j} = \tilde{\lambda}_j(u) P_j J_r(\tilde{\lambda}_{j}(u) d_j), \\
	&P_{j} = P_{j+1} A_{j+1}^T, P_{N} = I, A_{N} = I,j = 0 ,...., N-1.
	\label{dR_ddj}
\end{aligned}
\end{equation}

Where $A_{j}$ is defined as in eq.(\ref{omega_t}), and $A_{N} = I$ corresponds to the definition for $j = N$. It can be observed that, compared to the continuous trajectory representations in \cite{cumulative-Bspline, SLICT2}, the proposed formulation yields a more concise form for Jacobian derivation, without requiring additional consideration of boundary conditions. $\frac{\partial h_{\omega}}{\partial \hat{d}_{R,j} }$ can be computed as follows:
\begin{equation}
	\begin{aligned}
		\frac{\partial h_{\omega}}{\partial \hat{d}_{R,j} } & = \frac{\partial w(t) } {\partial d_j} = \frac{\partial w^{(j+1)}}{\partial d_j}, j = 0,...,N-1.
	\end{aligned}
\label{hw_jacobian}
\end{equation}

%根据eq.(\ref{omega_t}), $\frac{\partial w^{(j+1)}}{\partial d_j}$可以通过以下方式进行计算：
According to eq.(\ref{omega_t}), $\frac{\partial w^{(j+1)}}{\partial d_j}$ can be computed as follows:
\begin{align*}
	\frac{\partial w^{(j+1)}}{\partial d_j} = P_j(\tilde{\lambda}_j(u) A_j^T \left \lfloor w^{(j)} \right \rfloor_{\times} J_r(-\tilde{\lambda}_j(u) d_j) + \dot{\tilde{\lambda}}_j(u) I)
\end{align*}

%$b_a, b_w,g$与$d_{R,j}$无关，因此都为0。对于$\frac{\partial h_B}{\partial \hat{d}_{p,j} }$,根据观测方程$h_B(\chi, n)$以及eq.(\ref{velocity_acc})可以计算对应的雅可比：
$b_a$, $b_w$, and $g$ are independent of $d_{R,j}$, thus their derivatives are all zero. For $\frac{\partial h_B}{\partial \hat{d}_{p,j} }$, the corresponding Jacobian can be computed based on the observation equation $h_B(\chi, n)$ and eq.(\ref{velocity_acc}) as follows:
%\begin{equation}
%	\begin{aligned}
%		&\frac{\partial h_a}{\partial \hat{d}_{p,j} } = \frac{\partial {^{G}_{I}R(t)^{T}} a(t)}{\partial \hat{d}_{p,j} } = {^{G}_{I}R(t)^{T}} \ddot{\tilde{\lambda}}_{j}(u) I, \\
%		&\frac{\partial h_w}{\partial \hat{d}_{p,j} } = 0, j = 0,...,N-1.
%	\end{aligned}
%\end{equation}

\begin{equation}
	\begin{aligned}
		\frac{\partial h_a}{\partial \hat{d}_{p,j} } = {^{G}_{I}R(t)^{T}} \ddot{\tilde{\lambda}}_{j}(u) I, \frac{\partial h_w}{\partial \hat{d}_{p,j} } = 0, j = 0,...,N-1.
	\end{aligned}
\end{equation}

$b_a$, $b_w$, and $g$ are independent of $d_{p,j}$, thus their derivatives are all zero. Regarding $\frac{\partial h_B}{\partial \hat{\chi}_{I}}$, it can be computed as follows:
\begin{align*}
	&\frac{\partial h_a}{\partial \hat{\chi}_I} = \begin{bmatrix}
		0& I & -{^{G}_{I}R(t)^{T}} \left \lfloor \hat{g} \right \rfloor_{\times} B_{\hat{g}}
	\end{bmatrix}, \frac{\partial h_w}{\partial \hat{\chi}_I} = \begin{bmatrix}
		I& 0& 0
	\end{bmatrix}.
\end{align*}

Among these, $\frac{\partial h_B(\bar{\chi}, 0)}{\partial d_{R,j}}$ and $\frac{\partial h_B(\bar{\chi},0)}{\partial d_{p,j}}$ are related to the segment in which the historical state resides. The corresponding Jacobian only has non-zero values at the control points to be optimized, while the remaining control points in that segment are static and do not participate in the Jacobian computation.
$h_I$ is only related to the IMU state, therefore its Jacobian with respect to the state is:
\begin{align*}
	\frac{\partial h_I}{\partial \hat{\chi}} = \begin{bmatrix}
		0_{3 \times 24} & I_3 & 0_3 & 0_{3\times 2} \\
		0_{3 \times 24} & 0_3 & I_3 & 0_{3\times 2} \\
		0_{2 \times 24} & 0_{2 \times 3} & 0_{2 \times 3} & -\frac{B_{g_k}^T \left \lfloor g_k \right \rfloor_{\times} \left \lfloor g_k \right \rfloor_{\times} B_{g_k} }{||g_k||^2}
	\end{bmatrix}
\end{align*}

\begin{figure}[htbp]
	\centerline{\includegraphics[width=0.8\columnwidth ]{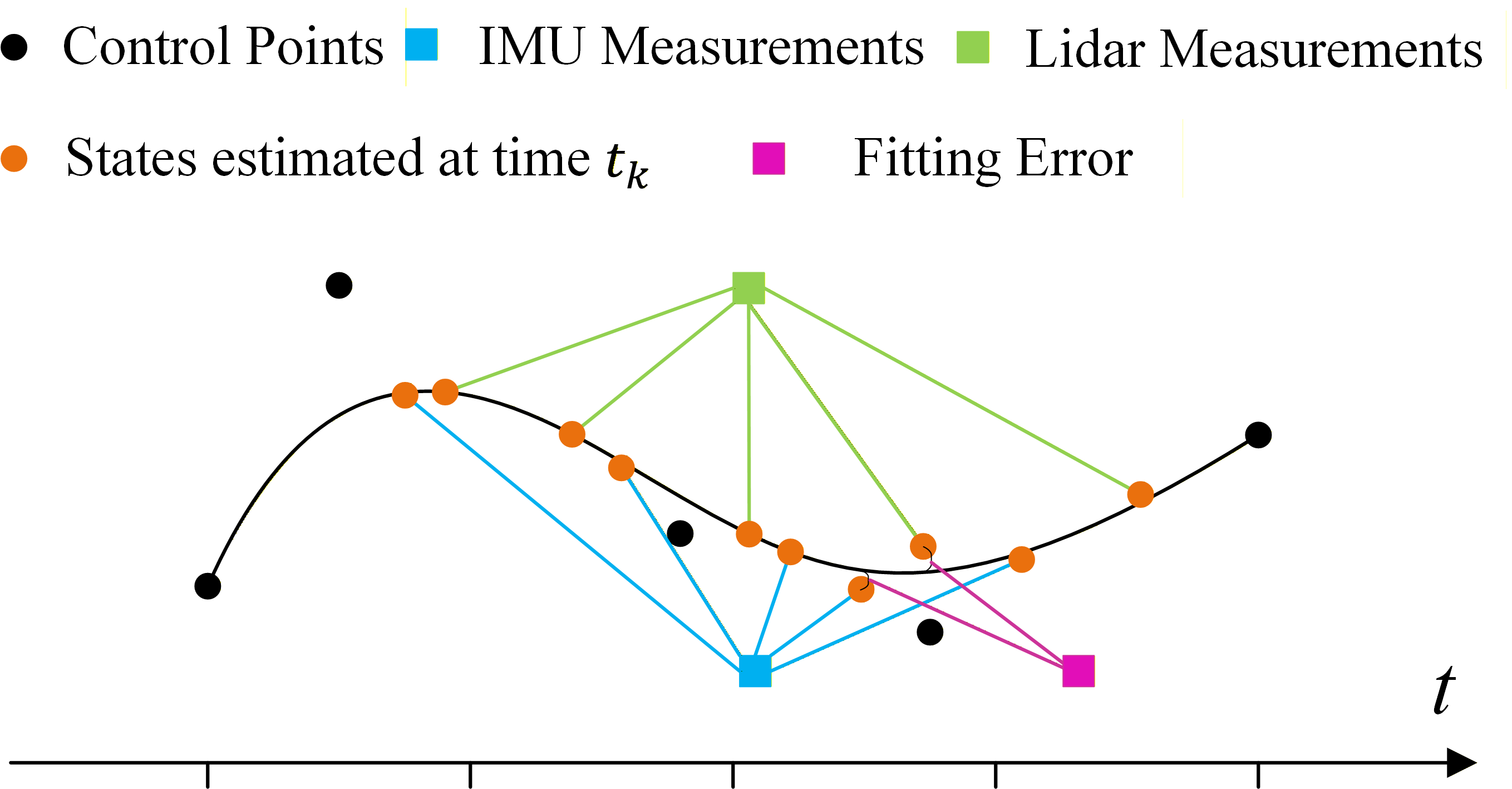}}
	\caption{Schematic diagram of representation error of continuous time spline curve.}
	\label{fitting_error}
\end{figure}

\textbf{Fitting error}:
In practical applications, to balance efficiency and representation accuracy, high-order splines are not employed, and it cannot be guaranteed that the poses at all discrete time instants $t_i, i = 1,...,n$ lie exactly on the continuous trajectory. Consequently, representation errors are inevitably introduced, as illustrated in Fig.\ref{fitting_error}, which subsequently affect the accuracy of pose estimation. Notably, this issue has not been addressed in existing continuous-time odometry works \cite{SLICT2, CLINS, CT-ICP, RESPLE}.

For the affected interval $[t_{k-N+2}, t_{k+1}]$, $\delta_{T}$ is estimated based on the differences between: (i) the optimally estimated poses over $[t_{k-N+2}, t_{k}]$ and the IMU forward-propagated poses over $[t_{k}, t_{k+1}]$ \cite{FAST-LIO2, FAST-LIVO2}, and (ii) the poses obtained via continuous trajectory interpolation at the corresponding timestamps. The representation error is generally small, and we assume it follows the distribution $\delta_{T} \sim \mathcal{N}(0, \Sigma_{T})$. In practice, rotation and translation are interpolated separately, leading to the respective assumptions $\delta_{R} \sim \mathcal{N}(0, \Sigma_{R})$ and $\delta_{p} \sim \mathcal{N}(0, \Sigma_{p})$. The specific calculation method is as follows:
\begin{equation}
\begin{aligned}
	\delta \theta_i &= {^{G}_{I_i}\bar{R}} \boxminus R(t_i), t_i \in [t_{k-N+2}, t_{k}], i = 1,...,n. \\
	\delta \theta_j &= {^{G}_{I_j}\hat{R}} \boxminus R(t_j), t_i \in [t_{k}, t_{k+1}], j = 1,...,s. \\
	\Sigma_{R} &= \frac{1}{N} \sum_{m = 1}^{N} \delta \theta_{m} \delta \theta_{m}^T, N = n+s.
\end{aligned}
\label{fitting_error_rotation}
\end{equation}

The above equation provides an estimate of the rotation representation error. For the translation representation error, a similar approach can be adopted.
\begin{equation}
\begin{aligned}
	\delta p_i &= {^{G}\bar{p}_{I_i}} \boxminus p(t_i), t_i \in [t_{k-N+2}, t_{k}], i = 1,...,n. \\
	\delta p_j &= {^{G}\hat{p}_{I_j}} \boxminus p(t_j), t_j \in [t_{k}, t_{k+1}], j = 1,...,s. \\
	\Sigma_{p} &= \frac{1}{N} \sum_{m = 1}^{N} \delta p_{m} \delta p_{m}^T, N = n+s.
\end{aligned}
\label{fitting_error_position}
\end{equation}

\subsection{Lidar Observation With Uncertainty}

\subsubsection{\textbf{Probalistic point to stable feature projection}}

Inspired by \cite{voxel_map}, we leverage point cloud uncertainty modeling \cite{voxel_map, CTA-LO, lio-uncertainty, LOG-LIO2} to characterize feature uncertainty, further improving registration accuracy and efficiency. However, unlike \cite{voxel_map} which focuses solely on planar features, we consider stable environmental structures more broadly. This is motivated by the fact that sensor placement or cluttered environments may not always provide sufficient and stable planar features, potentially degrading localization accuracy.

\textbf{Point Uncertainty}: 
%Adopting the point uncertainty model from \cite{voxel_map, pixel-cali}, we have:
%\begin{align*}
%	{^{L}p^{gt}_{i}} &= d^{gt}_{i} w^{gt}_{i} = (d + \delta_{d_i})(w_i \boxplus_{s^{2}} \delta_{w_i}) \\
%	& \simeq d_i w_i + \delta_{d_i} w_i- d_i \left \lfloor w_i \right \rfloor_{\times} N(w_i) \delta_{w_i} \\
%	& = {^{L}p_i} + \delta_{p_i}, \delta_{p_i} \sim N(0, {\textstyle \sum_{{^{L}p_i}}})
%\end{align*}
%
%where $\delta_{d_i} \sim N(0, {\textstyle \sum_{d_i}})$ be the ranging error, $\delta_{w_i} \sim N(0_{2 \times 1}, \textstyle \sum_{w_i})$ be the measurement noise in the tangent plane of $w_i$. $N(w_i) = [N_1, N_2] \in R^{3 \times 2} $ is an orthonormal basis of tangent space at $w_i$. $d_i w_i$ represents the LiDAR point measurement ${^{L}p_i}$, and $\delta_{p_i}$ denotes the corresponding measurement noise, with its specific form given as:
%\begin{align*}
%	{\textstyle \sum_{{^{L}p_i}}} = A_i \begin{bmatrix}
%	{\textstyle \sum_{d_i}} & 0_{1 \times 2} \\
%	0_{2 \times 1} & {\textstyle \sum_{w_i}}
%	\end{bmatrix}_{3 \times 3} A_i^{T}
%\end{align*}
%
%where $A_i = \begin{bmatrix} w_i & - d_i \left \lfloor w_i \right \rfloor_{\times} N_{w_i} \end{bmatrix}$. 
The uncertainty of the points is quantified using the method described in \cite{voxel_map, pixel-cali}. During pose estimation, points are projected into the global coordinate frame using the continuous-time representation: ${^{G}p_{i}} = {^{G}_{I}R(t)}({^{I}_{L}R} {^{L}p_i} + {^{I}t_{L}}) + {^{G}p(t)}$, which introduces additional localization uncertainty \cite{voxel_map}. The corresponding uncertainty of ${^{G}p_{i}}$ is given by:
\begin{small}
\begin{equation}
	\begin{aligned}
		&{\textstyle \sum_{^{G}p_i}} = {^{G}_{I}R(t)} {^{I}_{L}R} {\textstyle \sum_{^{L}p_{i}}} ({^{G}_{I}R(t)} {^{I}_{L}R})^T + {\textstyle \sum_{t}} \\
		&+ {^{G}_{I}R(t)} \left \lfloor{^{I}_{L}R} {^{L}p_i} + {^{I}t_{L}}\right \rfloor_{\times} {\textstyle \sum_{R}} \left \lfloor {^{I}_{L}R} {^{L}p_i} + {^{I}t_{L}} \right \rfloor_{\times}^{T} {^{G}_{I}R(t)}^{T} 
	\end{aligned}
	\label{point_cov_world}
\end{equation}
\end{small}
where ${\textstyle \sum_{t}}$ corresponds to the uncertainty of ${^{G}p(t)}$, and ${\textstyle \sum_{R}}$ corresponds to the uncertainty of ${^{G}_{I}R(t)}$.

\textbf{Plane Uncertainty}: For points ${^{G}p_i}, i = 1,...,N$ within a voxel, we can obtain:
\begin{align*}
	q = \frac{1}{N} \sum_{i = 1}^{N} {^{G}p_i}, {\textstyle \sum_{V}} = \frac{1}{N} \sum_{i = 1}^{N-1}({^{G}p_i} - q)({^{G}p_i} - q)^{T}.
\end{align*}

Let $q$ denote the mean of the point cloud distribution within the voxel, and ${\textstyle \sum_{V}}$ denote the corresponding covariance matrix. The eigendecomposition of this covariance matrix yields: $U\Lambda U^{T} = {\textstyle \sum_{V}}$, where $\Lambda = diag(\lambda_1, \lambda_2, \lambda_3)$ and $U = [u_1, u_2, u_3]$ represent the eigenvalues and eigenvectors, respectively. Due to the consideration of both LiDAR point measurement uncertainty and pose uncertainty, these uncertainties propagate to the eigenvalues, eigenvectors, and mean \cite{voxel_map}, yielding:
\begin{align*}
	 [\lambda^{gt}, u^{gt}, q^{gt}]^T &= f({^{G}p_1} + \delta_{^{G}p_1}, ..., {^{G}p_N + \delta_{^{G}p_N}}) \\
	 & \approx [\bar{\lambda}, \bar{u}, q]^T + \sum_{i = 1}^{N} \frac{\partial f}{\partial {^{G}p_i} } \delta_{^{G}p_i}.
\end{align*}

The corresponding covariance can be computed as ${\textstyle \sum_{\lambda, u, q}} = \sum_{i = 1}^{N} \frac{\partial f}{\partial {^{G}p_i} } {\textstyle \sum_{^{G}p_{i}}} (\frac{\partial f}{\partial {^{G}p_i}})^T$, 
where $\frac{\partial f}{\partial {^{G}p_i}} = [\frac{\partial \lambda_k}{\partial  {^{G}p_i} }, \frac{\partial u_k}{\partial {^{G}p_i}}, \frac{\partial q}{\partial {^{G}p_i}}]^T$. In this work, the uncertainty of the mean is not considered. 

%The uncertainties of the eigenvectors are computed as follows \cite{BALM}:
%\begin{align*}
%	&\frac{\partial u_k}{\partial p_i} = \begin{bmatrix}
%		\frac{\partial U e_k}{\partial x_i} & \frac{\partial U e_k}{\partial y_i} & \frac{\partial U e_k}{\partial z_i} 
%		\end{bmatrix} = U \begin{bmatrix}
%			F_{1, k}^{p_i} \\
%			F_{2, k}^{p_i} \\
%			F_{3, k}^{p_i}
%		\end{bmatrix}, \\
%		& F_{m, k}^{p_i} = \left\{\begin{matrix}
%		\frac{(^{G}p_i - \bar{p})^T}{N(\lambda_k - \lambda_m)} (u_m u_k^T + u_k u_m^T ), m \neq k\\
%		0, m = k.
%		\end{matrix}\right.
%\end{align*}
%\begin{align*}
%	\frac{\partial q}{\partial p_i} = diag(\frac{1}{N}, \frac{1}{N}, \frac{1}{N})
%\end{align*}
%
%where $e_k$ is a $3 \times 1$ column vector with the $k$-th element equal to 1 and all other elements equal to 0, and $N$ denotes the number of points.

\begin{figure}[htbp]
	\centerline{\includegraphics[width=0.8\columnwidth ]{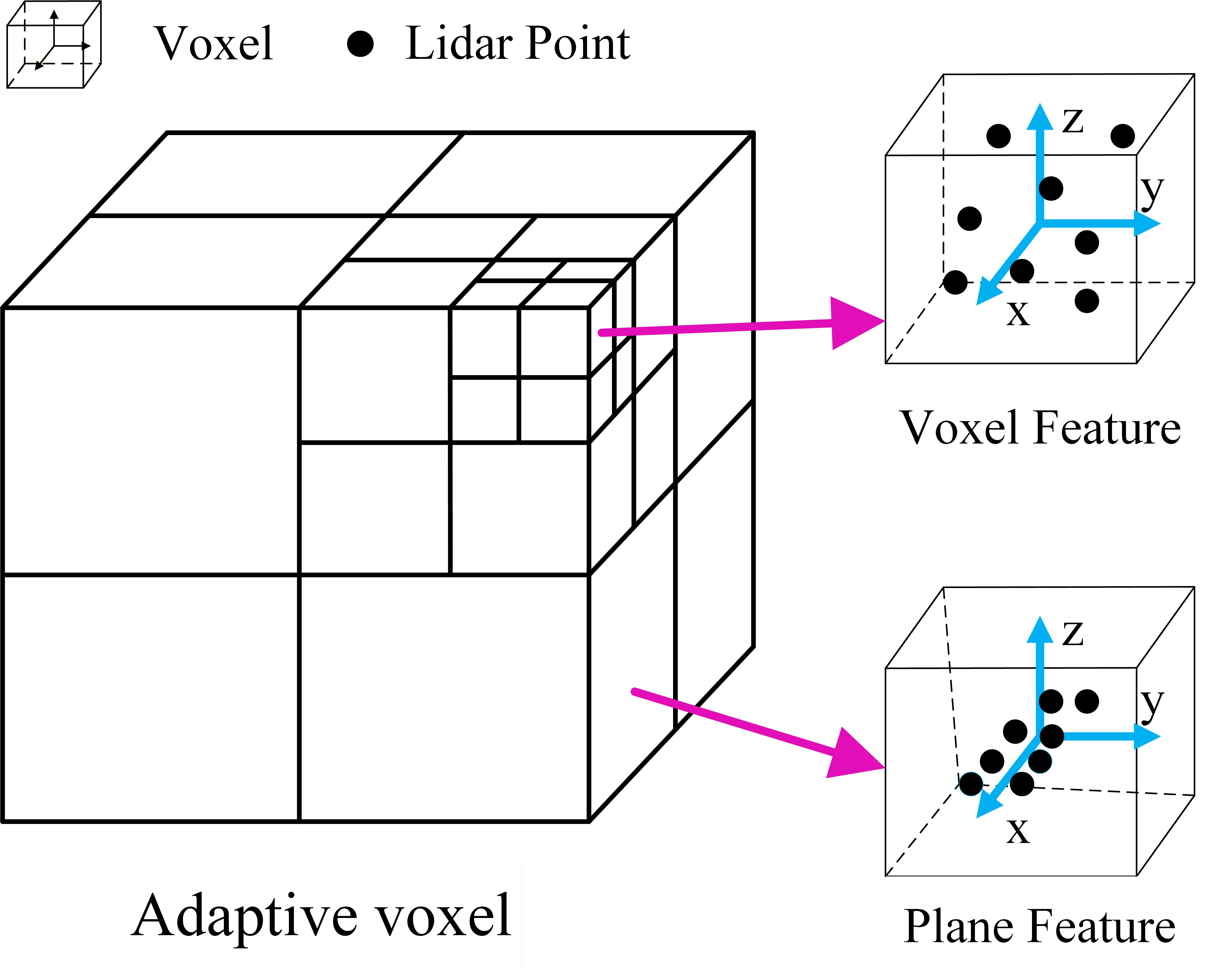}}
	\caption{Voxel representation and decomposition diagram.}
	\label{voxel_feature}
\end{figure}

\textbf{Point matching with feature uncertainty}: 
During point cloud registration, relying on a single feature type may lead to performance degradation when the corresponding feature cannot be reliably detected \cite{voxel_map}. Therefore, we employ multiple feature types in this work. Considering the sparsity of LiDAR point clouds, we adopt both plane features and voxel features, denoted as $f_{p}$ and $f_{v}$, respectively. Unlike previous works that primarily employ ICP-style approaches \cite{x-icp, FAST-LIO2, BALM}, our uncertainty modeling enables precise characterization of feature reliability, leading to more robust registration performance. The specific feature representations are illustrated in Fig.~(\ref{voxel_feature}).

The voxel construction follows a coarse-to-fine strategy similar to \cite{voxel_map}. Specifically, when the point cloud is sparse, a coarse voxel representation is established. As the point cloud becomes denser, an adaptive refinement strategy is employed to construct finer voxels. (Note: When a plane feature can be extracted from a voxel, further subdivision is not performed; otherwise, the voxel is subdivided until either stable features can be extracted or the maximum subdivision depth is reached.) Voxels that are leaf nodes and do not contain plane features are designated as voxel features.
After constructing the voxels, different types of observation equations are formulated based on the feature type.

\textit{Point-to-plane}: The corresponding observation equation is formulated as follows:
\begin{small}
\begin{equation}
	\begin{aligned}
		&h_{L}(\chi, n) = u_1^T ({^{G}p_i} - q)\\
		&= (\hat{u}_1 \boxplus \delta_{u_1})^{T}[ (\hat{T}(t_i) \boxplus \delta_{T}) {^{I}T_{L}} ({^{L}p_i}+ \delta_{^{L}p_i}) - q - \delta_q] \\
		&\simeq \underbrace{ \hat{u}_1^T (\hat{T}(t_i) {^{I}T_{L}} {^{L}\hat{p}_i} - \hat{q})}_{h_{L,r}(\hat{\chi}, 0)} + H_{L}\delta \chi \\
		&+ \underbrace{ J_{u_1} \delta_{u_1} + J_{q} \delta_{q} + J_{R(t_i)}\delta_{\theta_{R(t_i)}} + J_{p(t_i)} \delta_{p(t_i)} + J_{^{L}p_{i}} \delta_{^{L}p_i}}_{n_r}
	\end{aligned}
	\label{point_to_plane}
\end{equation}
\end{small}

where $h_{L,r}(\hat{\chi}, 0)$ denotes the residual, and $n_r$ represents the noise. $\hat{T}(t_i) \in SE(3)$ denotes the pose computed from the continuous trajectory at time ${^{L}\hat{p}_i}$, and $\delta_{T}$ represents the fittting error of the continuous trajectory. $H_{L, i}$ denotes the Jacobian matrix evaluated at $\hat{\chi}$, which can be computed as follows:
\begin{equation}
	\begin{aligned}
		H_{L, i} &= \lim_{\delta \chi \to 0} \frac{h_{L,r}(\hat{\chi} \boxplus \delta \chi, 0) \boxminus h_{L,r}(\hat{\chi}, 0)}{\delta \chi} \\
		& = \begin{bmatrix}
		\frac{\partial h_{L,r}}{\partial d_{R,j} } & \frac{\partial 	h_{L,r}}{\partial d_{p, j} } & \frac{\partial h_{L,r}}{\partial \chi_{I}}
		\end{bmatrix}, j = 0,...,N-1.
	\end{aligned}
\label{obs_jacobian}
\end{equation}

%其中，$\frac{\partial h_{L,r}}{\partial \chi_{I}} = [0_{1 \times 8}]$，关于样条曲线状态量的雅可比矩阵可以通过链式法则进行求导,即有：
where $\frac{\partial h_{L,r}}{\partial \chi_{I}} = [0_{1 \times 8}]$. The Jacobian matrix with respect to the spline state variables can be derived using the chain rule as follows:
\begin{align}
	\frac{\partial h_{L,r}}{\partial d_{R,j} } = \frac{\partial h_{L,r}}{\partial R(t_i)} \frac{\partial R(t_i)}{\partial d_{R,j}}, \frac{\partial h_{L,r}}{\partial d_{p, j} } = \frac{\partial h_{L,r}}{\partial p(t_i)} \frac{\partial p(t_i)}{\partial d_{p, j}}
	\label{partial_chain}
\end{align}

%根据观测方程，可以得到：
Based on the observation equation, we can obtain:
\begin{align*}
	\frac{\partial h_{L,r}}{\partial R(t_i)} = -\hat{u}_1^T {\hat{R}(t_i)} \left \lfloor {^{I}T_{L}} {^{L}p_i}  \right \rfloor_{\times}, \frac{\partial h_{L,r}}{\partial p(t_i)} =  \hat{u}_1^T.
\end{align*}

%$\frac{\partial R(t_i)}{\partial d_{R,j} }$可以通过eq.(\ref{dR_ddj})计算，$\frac{\partial p(t_i)}{\partial d_{p,j}}$可以通过以下的方式进行计算：
$\frac{\partial R(t_i)}{\partial d_{R,j} }$ can be computed via eq.(\ref{dR_ddj}), and $\frac{\partial p(t_i)}{\partial d_{p,j}}$ can be computed as follows:
\begin{equation}
\begin{aligned}
	\frac{\partial p(t_i)}{\partial d_{p,j}} = \tilde{\lambda}_{j}(u)I, j = 0,...,N-1;
	\label{dp_ddj}
\end{aligned}
\end{equation}

Based on the formulation of the observation equation, which implies that each dimension of $h_{L}(\chi, n)$ follows an approximately zero-mean Gaussian distribution:
\begin{align*}
	& h_{L, i}(\chi, n) \sim N(h_{L,i}(\hat{\chi}, 0), \textstyle \sum_{n_i}), \textstyle \sum_{n_i} = J_{n_i} \textstyle \sum J_{n_i}^T \\
	&{\textstyle \sum }= diag({\textstyle \sum_{u_1}}, {\textstyle \sum_{q}},{\textstyle \sum_{^{L}p_i}},{\textstyle \sum_{R(t_i)}},{\textstyle \sum_{p(t_i)}} )
\end{align*}

where $J_{n_1}$ is computed as follows:
\begin{align*}
	&J_{n_i} = [J_{u_1}, J_{q}, J_{^{L}p_i}, J_{R(t_i)}, J_{p(t_i)}] \\
	&J_{u_1} = ( {\hat{T}(t_i)} {^{I}T_{L}} {^{L}\hat{p}_i} - q)^T, J_{q} = - \hat{u}_1^T, J_{^{L}p_i} =  \hat{u}_1^T \hat{R}(t_i) {^{I}_{L}R}, \\
	&J_{R(t_i)} = -\hat{u}_1^T {\hat{R}(t_i)} \left \lfloor {^{I}T_{L}} {^{L}\hat{p}_i}  \right \rfloor_{\times}, J_{p(t_i)} = \hat{u}_1^T.
\end{align*}

\textit{Point-to-voxel}: When points do not lie within planar regions, voxel features are further utilized for registration. The local structure within a voxel is characterized by the three eigenvalues and eigenvectors of the spatial point cloud distribution, and the observation equation is formulated by projecting points onto this distribution. Different from \cite{HierD-lio, 3DMNDT, NDT}, by modeling both LiDAR point uncertainty and pose uncertainty, the three eigenvectors and eigenvalues are treated as random variables.
%as illustrated in Fig.\ref{local_structure}. This enables the use of additional voxel features for registration in the absence of planar structures, thereby enhancing system robustness.
\begin{small}
\begin{equation}
	\begin{aligned}
		&h_{L}(\chi, n) = \bar{\Lambda}_i^{-\frac{1}{2}} U^{T}_i ({^{G}p_i} - q) = \begin{bmatrix}
			\bar{\lambda}_1^{-\frac{1}{2}}u_{1}^{T} ({^{G}p_i} - q) \\
			\bar{\lambda}_2^{-\frac{1}{2}}u_{2}^{T} ({^{G}p_i} - q)\\
			\bar{\lambda}_3^{-\frac{1}{2}}u_{3}^{T} ({^{G}p_i} - q)
		\end{bmatrix} = \\
		&\begin{bmatrix}
			k_1 (\hat{u}_1 \boxplus \delta_{u_1})^{T}[ (\hat{T}(t_i) \boxplus \delta_{T_i}) {^{I}T_{L}} ({^{L}\hat{p}_i}+ \delta_{^{L}p_i}) - \hat{q} - \delta_{q}] \\
			k_2 (\hat{u}_1 \boxplus \delta_{u_2})^{T}[ 	(\hat{T}(t_i) \boxplus \delta_{T_i}) {^{I}T_{L}} ({^{L}\hat{p}_i}+ \delta_{^{L}p_i}) - \hat{q} - \delta_{q}] \\
			k_3 (\hat{u}_3 \boxplus \delta_{u_3})^{T}[ (\hat{T}(t_i) \boxplus \delta_{T_i}) {^{I}T_{L}} ({^{L}\hat{p}_i}+ \delta_{^{L}p_i}) - \hat{q} - \delta_{q}]
		\end{bmatrix}
		\label{point_to_voxel}
	\end{aligned}
\end{equation}
\end{small}
where ${\textstyle \sum_{V}} = U \Lambda U^T = U \Lambda^{\frac{1}{2}} \Lambda^{\frac{1}{2}} U^T = (U \Lambda^{-\frac{1}{2}} \Lambda^{-\frac{1}{2}} U^T)^{-1}$, $k_i = \bar{\lambda}_i^{-\frac{1}{2}}, i =1,2,3$. The computation of the Jacobian matrix and covariance for voxel features is similar to that of planar features. The main difference lies in that voxel features require three separate computations, each additionally weighted by the corresponding eigenvalue as coefficients. We omit further detailed derivation here for brevity.

%Taking one row from the above observation equation as an example, we can obtain:
%\begin{align*}
%	&h_{L,j}(\chi_k, n) = \\
%	&\bar{\lambda}_j^{-\frac{1}{2}} (\hat{u}_j \boxplus \delta_{u_1})^{T}[ (\hat{T}(t_i) \boxplus \delta_{T_i}) {^{I}T_{L}} ({^{L}\hat{p}_i}+ \delta_{^{L}p_i}) - \hat{q} - \delta_{q}] \\
%	&\simeq \underbrace{ \bar{\lambda}_j^{-\frac{1}{2}} \hat{u}_j^T (\hat{T}(t_i) {^{I}T_{L}} {^{L}\hat{p}_i} - \hat{q})}_{h_{L,r}(\hat{\chi}, 0)} + H_{L}\delta \chi \\
%	&+ \underbrace{ J_{u_j} \delta_{u_j} + J_{q} \delta_{q} + J_{R(t_i)}\delta_{\theta_{R(t_i)}} + J_{p(t_i)} \delta_{p(t_i)} + J_{^{L}p_{i}} \delta_{^{L}p_i}}_{n_r}.
%\end{align*}

%\begin{figure}[htbp]
%	\centerline{\includegraphics[width=0.6\columnwidth ]{pictures/local_structure.png}}
%	\caption{The diagram of lidar to voxel projection.}
%	\label{local_structure}
%\end{figure}

\textit{Feature filtering}: 
Considering the sparsity of LiDAR scans, accurately characterizing the local structure of a voxel typically requires observations accumulated over multiple scans. Therefore, when projecting points based on feature types, a filtering mechanism based on the $3\sigma$ principle is applied. Taking point-to-plane matching as an example, with the filtering mechanism incorporated, this can be expressed as:
\begin{equation}
	\begin{aligned}
		h_{L}(\chi_k, n) &= s u_1^T ({^{G}p_i} - q), \\
		s &= \left\{\begin{matrix}
			1 & if \, d_{i} \in U(q, 3 \sqrt{\textstyle \sum_{n_j}} ), \\
			0 & otherwise
		\end{matrix}\right. 
	\end{aligned}
\label{point_filter}
\end{equation}
where $U(q, 3 \sqrt{\textstyle \sum_{n_j}} )$ denotes the $3\sigma$ neighborhood of $q$, and $d_{i}$ is the distance to the corresponding plane.

\iffalse
具体表示为：
\begin{align*}
	d_{i} &= u_{1}^{T} ({^{G}p_i} - q) \\
	&= (\hat{u}_{1} + \delta_{u_1})^T ( {^{G}\hat{p}_i} + \delta_{^{G}p_i}  - \hat{q} - \delta_{q}) \\
	&\simeq \hat{u}_{1}^T ({^{G}\hat{p}_i} - \hat{q}) + J_{u_1}\delta_{u_1} + J_{{^{G}p_i}}\delta_{^{G}p_i}  + J_{q} \delta_{q}
\end{align*}

which implies that $d_{i} \sim N(0, {\textstyle \sum_{n_i}} )$, where,
\begin{align*}
	{\textstyle \sum_{n_i}} &= J_{n_i} {\textstyle \sum_{u_1, q, {^{G}p_i}}} J_{n_i}^T\\
	J_{n_i} &= \begin{bmatrix}
		J_{u_i} & J_{q} & J_{^{G}p_i}
	\end{bmatrix} \\
	& = \begin{bmatrix}
		(^{G}\hat{p} - \hat{q})^T & -\hat{u}_1^T & \hat{u}_1^T
	\end{bmatrix} \\
	{\textstyle \sum_{u_1, q, {^{G}p_i}}} &= \begin{bmatrix}
		\sum_{u_1} & 0& 0\\
		0 & \sum_{q} & 0\\
		0 & 0 & \sum_{^{G}p_i}
	\end{bmatrix}
\end{align*}
\fi

For voxel features, the judgment is performed on all three planes. A match is considered valid only when the $3\sigma$ principle is satisfied for all three planes simultaneously.

\subsubsection{\textbf{re-estimated policy}}
\begin{figure}[htbp]
	\centerline{\includegraphics[width=0.8\columnwidth ]{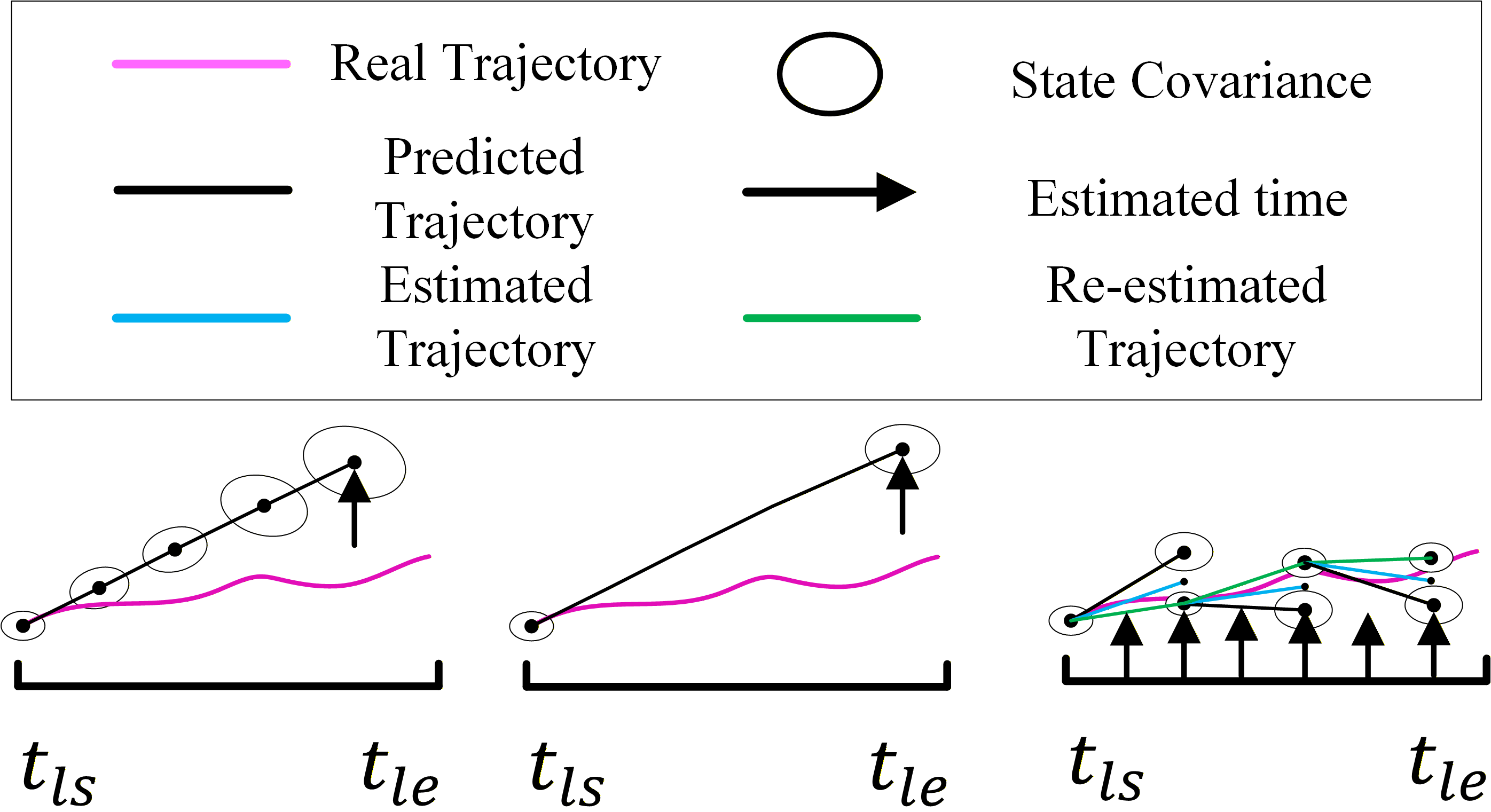}}
	\caption{The schematic diagram illustrates three prediction and estimation strategies, with the rightmost strategy being the prediction and re-estimation strategy used in this paper.}
	\label{reestimated_policy}
\end{figure}
In this work, state prediction is performed under a constant velocity assumption. When the prediction interval $\delta t$ is short, multiple predictions occur within the time interval $[t_{ls}, t_{le}]$ of a single LiDAR scan, causing the uncertainty of control points to accumulate rapidly. Consequently, the subsequent posterior estimation, relying on only a few iterations, may not yield satisfactory results. Conversely, when the prediction interval is long, the simple motion assumption fails to accurately characterize the motion within the interval, leading to poor initial guesses and subsequently degraded estimation performance.

To address this issue, inspired by \cite{point-lio}, we propose a re-estimation policy. Specifically, a relatively short prediction interval is adopted, and a maximum number of re-estimations (also the maximum sampling times) is set. Once the number of points in the current interval exceeds a predefined threshold, fixed-number sampling is initiated, and the above process is repeated until either the maximum number of re-estimations is reached or all points in the current interval have been processed. This strategy significantly reduces the number of observations involved in computing the Kalman gain per estimation, thereby improving computational efficiency. Moreover, multiple iterative estimations are performed within a short interval (using the previous estimate as prior), enabling the estimated state to better align with the actual robot motion. This approach achieves remarkable performance and robustness, as illustrated in Fig.(\ref{reestimated_policy}).

Taking the prediction interval $\delta t$ as an example, upon the arrival of a new LiDAR scan, the state is propagated once. Let the start and end times of this propagation be denoted as $t_{ps}$ and $t_{pe}$, respectively. When the number of points within the corresponding interval $[t_{ps}, t_{pe}]$ exceeds the threshold $N_{thre}$, a sample-estimate-resample process is initiated. Assuming the final number of sampling iterations is $k$, i.e., the point cloud is divided into $k$ segments, the efficiency improvement can be analyzed in terms of computational complexity.

Since residual and Jacobian computations scale linearly with the number of points, the computational efficiency in this part remains largely unaffected. The primary difference lies in the Kalman gain computation and covariance matrix update. Based on the Kalman gain $K \in \mathbb{R}^{m \times n}$ in eq.(\ref{delta_x}), the covariance $P \in \mathbb{R}^{n \times n}$, and the computational complexity analysis in \cite{mckf}, we obtain:
\begin{equation}
	\begin{aligned}
		S_{K} &= 6 n^{2}m + 2nm^{2} + n^{2} - 4mn + O(m^{3}) + O(n^{3}) \\
		S_{P} &= 2 n^{3} + 2n^{2}m - n^{2}
	\end{aligned}
\end{equation}

where $S_{P}$ denotes the time complexity of the covariance update, and $S_{K}$ denotes the time complexity of computing the Kalman gain (a conservative estimate; in practice, since $R$ is a diagonal matrix, its inversion can be simplified to $O(m)$). $n$ represents the state dimension, and $m$ represents the observation dimension. Thus, the re-estimation policy effectively decomposes an $m$-dimensional problem into $k$ subproblems of dimension $m/k$, thereby improving computational efficiency.

\subsection{State Update}
Based on the iterated extended Kalman filter (IEKF) framework, maximum a posteriori (MAP) estimation is performed iteratively according to the observation equation and kinematic equation:
\begin{equation}
	\begin{aligned}
		&\delta \chi_{k,i+1} = \\
		&K_{i}(z_k - h(\hat{\chi}_{k|k, i}, 0)) - (I- K_i H_i) (\hat{\chi}_{k|k, i} - \hat{\chi}_{k| k -1}). 
	\end{aligned}
\label{delta_x}
\end{equation}
where $K_i = (H_i^T R_{k}^{-1} H_i + P_{k|k}^{-1})^{-1} H_{i}^{T} R_{k}^{-1}$ is the Kalman gain, $z_k$ is the ideal observation set to $0$, and $\hat{\chi}_{k|k, i}$ denotes the state estimate at the $i$-th iteration. $H_{i} = [H_B^T, H_L^T ]^T$, $R_{k} = diag(\sum_{B}, \sum_{L})$. The iteration terminates when either the maximum number of iterations $N$ is reached or $||\delta \chi_{k,i+1}|| \leq \epsilon$, where $\epsilon$ is a sufficiently small positive threshold. Upon termination, the covariance is updated as $P_{k|k} = (I- K_i H_i) P_{k|k-1}$.
Based on the above discussion, the algorithmic framework is summarized in Algorithm \ref{ctvoxel_map}.
\begin{algorithm}[htbp]  
	\caption{ CTVoxelMap: Continuous-time state estimation with adaptive voxel map}  
	\label{ctvoxel_map}
	\begin{algorithmic}[1]
		\item[] \textbf{Input:} ${\Delta t}$: the knot's interval time of cubic B-spline; ${^{L}p_i^{t}}, i = 1,...,n$: the $k$-th frame measurements of lidar; $\{v_t, \omega_t \}$: the measurements of imu, $t \in [t_{k-j}, t_{k}]$, $j = 1, ...$ ; $\hat{\chi}_k$: the state to be estimated;  
		\item[] \textbf{Output:} $\bar{\chi}_k$ : the state estimated. 
		\Repeat
		\State Propagate the state once.
		\State Let $N_l$ denote the set of LiDAR points acquired within the propagation interval.
		\Repeat
		\If {$N_l > N_{thre}$}
		\State Devide $N_l$ into two parts, $N_{l1}$ and $N_{l2}$, let $N_l = N_{l1}$.
		\EndIf
		\State calculate the pose in $\{G\}$ of lidar points in $N_l$. 
		\State If the voxel map is not initialized, construct the adaptive voxel map(Fig.(\ref{voxel_feature})).
		\State Based on the $\chi_{k}$ and imu measurements, calculate the cubic Bspline fitting error.
		\State $i = 0$, $\chi_{k,i} = \hat{\chi}_k$.
		\For {$||\chi_{k,i+1} \boxminus \chi_{k,i}|| > \epsilon$ or $i < N$}
		\State Based on measurements and eq.(\ref{Bspline_observation}), calculate the residual and jacobian with respect to $\chi_{k, i}$.
		\State Based on voxel map, the fitting error and the observation formulation of features, calculate the residual and jacobian with respect to $\chi_{k,i}$.
		\State Calculate the $\delta \chi_{k, i}$ according to eq.(\ref{delta_x}).
		\State $\chi_{k, i+1} = \chi_{k,i} \boxplus \delta \chi_{k,i}$
		\State Recompute the pose of lidar points in $\{ G\}$ according to $\chi_{k, i+1}$ and update voxel map.
		\State Update the cubic Bspline fitting error.
		\State $i = i + 1$.
		\EndFor
		\State Get the estimated state $\bar{\chi}_{k}$. 
		\State $N_l = N_{l2}$, $\hat{\chi}_k = \bar{\chi}_{k}$.
		\Until {all points in this frame is processed or the max re-estimation times is reached.}
		\Until{all points is processed.} \\
		\Return $\bar{\chi}_{k}$.
	\end{algorithmic}  
\end{algorithm}

\section{Experiments}
The performance of the proposed algorithm is comprehensively evaluated through extensive experiments and detailed ablation studies on public datasets, supplemented by further real-robot validation on a wheel-legged robot.

\begin{figure*}[htbp]
	\centering
	\subfloat[M2UD\cite{M2UD}]{
		\includegraphics[height=3.35cm]{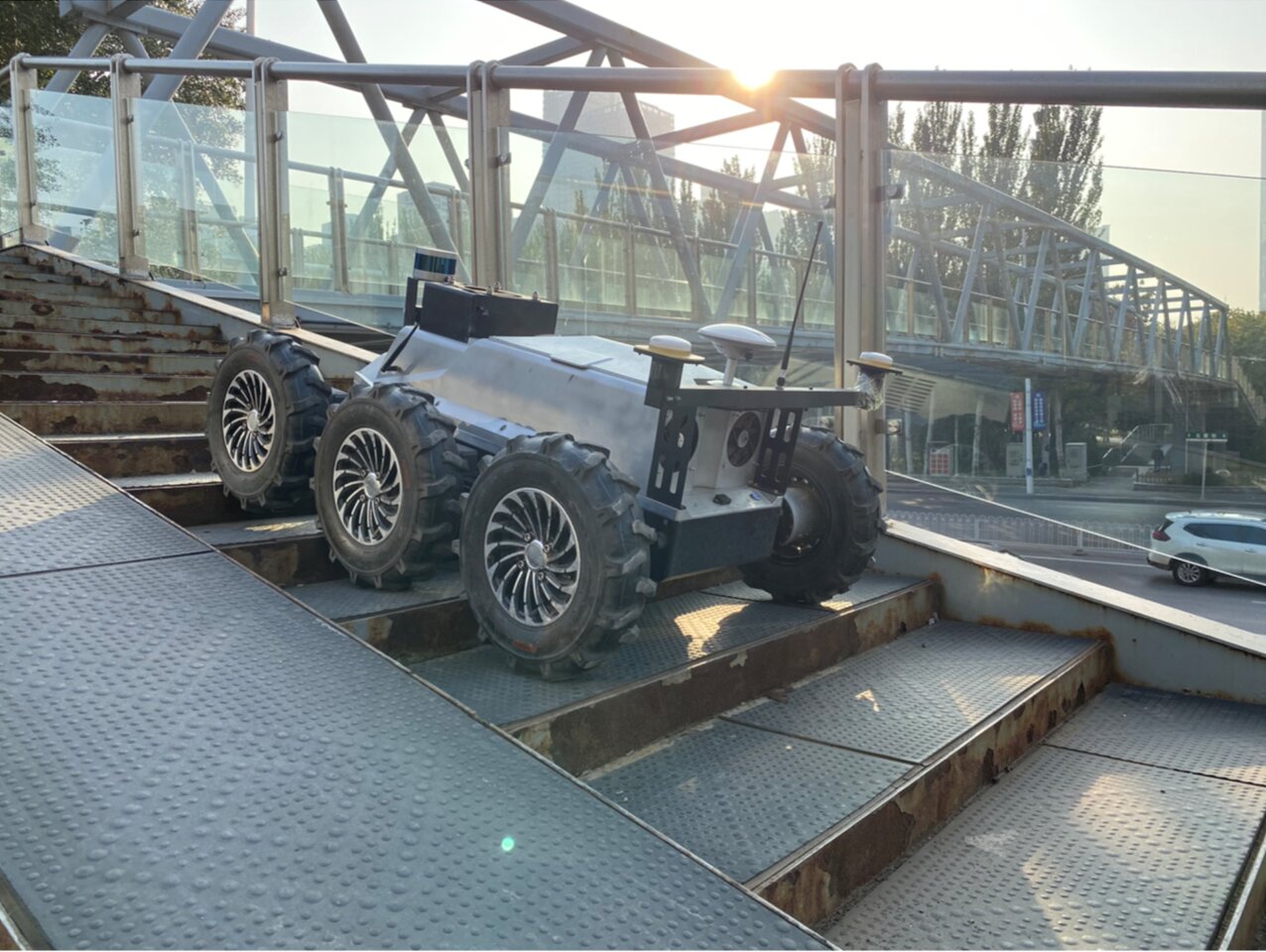}
	}
	\hfill
	\subfloat[MARS-LVIG\cite{MARS_LVIG}]{
		\includegraphics[height=3.35cm]{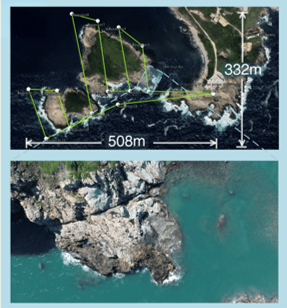}
	}
	\hfill
	\subfloat[Diter++\cite{Diter++}]{
		\includegraphics[height=3.35cm]{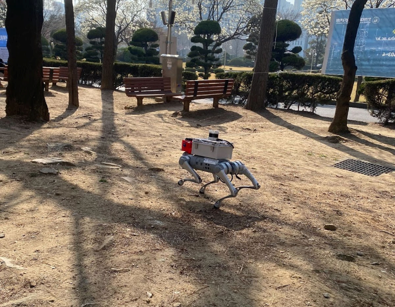}
	}
	\hfill
	\subfloat[MCD\cite{MCD}]{
		\includegraphics[height=3.35cm]{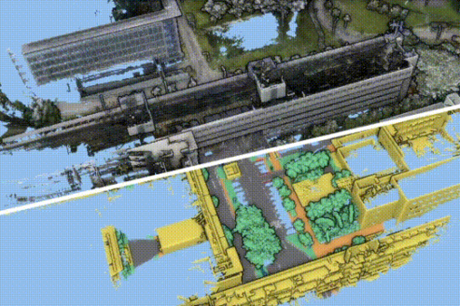}
	}
	\caption{The experiments are conducted on four public datasets with diverse platforms and environmental conditions. M2UD is collected using an all-terrain vehicle operating in staircases, forests, and playgrounds, featuring rugged terrain and low-speed motion. MARS-LVIG is captured by a quadrotor UAV flying in a downward-facing view at high speed. Diter++ is collected using a quadruped robot navigating complex terrain with significant vibrations and irregularities. MCD is acquired from a ground mobile robot operating in campus environments, characterized by relatively high-speed motion and a LiDAR sensor with a limited field of view.}
	\label{datasets}
\end{figure*}

\subsection{Experiment setup}
The datasets used in this work include M2UD \cite{M2UD}, MARS-LVIG \cite{MARS_LVIG}, Diter++\cite{Diter++} and MCD \cite{MCD}. These datasets encompass diverse indoor and outdoor scenarios and feature various robotic platforms, such as all-terrain vehicles,, unmanned aerial vehicles, quadruped and ground mobile robots. Detailed information is provided in Table.\ref{dataset} and Figure.\ref{datasets}.
\begin{table}[htbp]
	\centering
	\caption{Dataset for real-world}
	\begin{tabular}{M{1cm} M{3cm} M{3cm}}
		\hline
		\textbf{Dataset} & \textbf{Scenarios} & \textbf{Sensors(Adopted)} \\
		\hline
		M2UD\cite{M2UD} & urban, palza, parking, campus, ground robot, uneven-terrain & L(Velodyne VLP-16), I(Xsens MTi 300) \\
		\hline
		MARS LVIG \cite{MARS_LVIG}& airport, island, and valley, large-scale, fast, drone & L(AVIA), I(built-in)\\
		\hline
		Diter++\cite{Diter++}& quadruped, forest, and park, large-scale, rough-terrain & L(Ouster OS1 64/128), I(built-in)\\
		\hline
		MCD\cite{MCD} & large-scale urban, fast, ground vehicle  & L(Livox Mid 70), I(VN100)\\
		\hline
	\end{tabular}
\label{dataset}
\end{table}

For method comparison, the following algorithms are selected as baselines: FAST-LIO2 (F-LIO2) \cite{FAST-LIO2}, FAST-LIVO2 (F-LIVO2) \cite{FAST-LIVO2}, CLIC \cite{CLIC}, and RESPLE (R-LIO) \cite{RESPLE}. F-LIO2 is a widely adopted LiDAR-inertial odometry framework. F-LIVO2, an improved version of VoxelMap, achieves state-of-the-art performance and has garnered significant attention as a multi-sensor fusion odometry method; its LIO component is employed in the comparative experiments. Both CLIC and RESPLE are recent continuous-time odometry methods, with CLIC adopting an optimization-based framework and RESPLE utilizing an IEKF-based framework, both demonstrating competitive performance.

To quantify accuracy, trajectory estimates are interpolated at the end timestamps of LiDAR frames. The root mean square error (RMSE) and maximum value (MAX) of the absolute position error (APE) are then computed using the evo tool \cite{evo}. To mitigate the influence of random errors, each sequence in every dataset is run three times, and the resulting RMSE and MAX metrics are averaged to obtain the final values. 

Due to variations in platforms and LiDAR sensor types across different datasets, experimental parameters are adjusted accordingly. In the MCD dataset, the knot frequency is set to $80$Hz. For the M2UD and MARS-LVIG datasets, a knot frequency of $50$Hz achieves optimal accuracy and robustness. For the Diter++ dataset , the knot frequncy is set to $35$HZ. All algorithms are evaluated on the same hardware platform with the following specifications: CPU: i5-12400F, RAM: 16GB.
%In all experiments, the state covariance noise is maintained at an order of magnitude around $10^{-7}$, while the system noise is set to approximately $10^{-6}$ for old control points and $10^{-8}$ for new control points, ensuring favorable robustness and performance. These settings are consistently applied when adjusting the control point frequency.

\subsection{Performance comparison on real-world datasets}

\begin{table*}[htbp]
	\centering
	\caption{APE (MAX/ RMSE, meters) of methods on public dataset}
	\renewcommand{\arraystretch}{1.4} 
	\begin{tabular}{c c c c c c c c}
		\hline 
		\textbf{Dataset} & \textbf{Sequence} & \textbf{CLIC} & \textbf{F-LIO2} &  \textbf{R-LIO} & \textbf{F-LIVO2} & \makecell{\textbf{CT-VLO} \\ \textbf{(ours)}} & \makecell{\textbf{CT-VLIO} \\ \textbf{(ours)}} \rule{0pt}{15pt} \\
		\hline
		\multirow{6}{*}{MCD} & ntu-day-01 & 25.194 & 0.901 & $\mathbf{0.549}$& 3.681 / 1.276 & 2.639 / 1.100 & 2.904 / 1.444\\
		 & ntu-day-02 & 0.503 & 0.185 &0.188 & 0.439 / 0.148  & 0.421 / 0.167 & 0.355 / $\mathbf{0.144}$\\
		 & ntu-day-10 & 22.775 &1.975 &1.493 & 2.127 / 0.826 & 1.549 / $\mathbf{0.646}$& 2.295 / 0.955\\
		 & ntu-night-04 & 4.154 & 0.902 &0.416 & 1.028 / 0.455 & 1.063 / 0.385 & 0.901 / $\mathbf{0.310}$\\
		 & ntu-night-08 & 22.275 & 1.002 &0.940 & 2.133 / 0.868 & 1.983 / $\mathbf{0.789}$ & 2.242 / 1.132\\
		 & ntu-night-13 & 21.299 &1.288 &0.560 & 2.587 / 1.112 & 0.855 / $\mathbf{0.381}$& 2.452 / 1.229 \\
		\hline
		\multirow{4}{*}{M2UD} & plaza03 & $\times$ & $\mathbf{ 0.831 }$/ 0.436 & 1.733 / $\mathbf{0.308^{*}} $ & $\times$ & 2.419 / 0.644 & 1.435 / 0.619\\
		 & aggressive04 & $\times$ & $\times$ & 2.507 / 0.882  &$\times$ &  $\mathbf{2.107 / 0.690}$ & 2.118 / 0.715\\
		 & aggressive05 & $\times$ & $\mathbf{7.298}$ / 3.567 & 7.465 / 3.228 &$\times$ & 7.521 / $\mathbf{3.179 }$ & 7.548 / 3.243\\
		 & campus06 & $\times$ & 6.601 / 2.524 & 3.642 / 1.422 & 2.392 / 
		 1.256 & $\mathbf{2.292 / 1.123}$ & 3.011 / 1.461 \\
		\hline
		\multirow{3}{*}{MARS-LVIG}& HKisland02 & $\mathbf{3.475}$ / 1.973 & 3.851 / 2.241 & 4.503 / 2.177 & 3.948 / 2.028 &  4.562 / 2.065 & 4.232 / $\mathbf{1.909}$\\
		 & HKairport02 & 15.635 / 3.122 & 8.987 / 2.931 & 10.797 / 3.557 & 8.390 / 3.284 & 15.294 / 3.220 & $\mathbf{4.509 / 2.702}$  \\
		 & AMtown03 & $\times$ & 6.191 / 3.216 & 20.640 / 12.630 & 5.490 / 2.685 & 3.971 / 2.299 & $\mathbf{3.183 / 2.033}$ \\
		\hline
		\multirow{2}{*}{Diter++} & Park-in-day-cam-free & $\times$ & 0.939 / 0.348 & 0.825 / 0.416 & 3.194 / 1.011 & 0.581 / $\mathbf{ 0.214}$ & $\mathbf{0.573}$ / 0.252 \\
		 & Forest-new & 2.704 / 1.372 & 0.438 / $\mathbf{0.103}$ & $\mathbf{0.371}$ / 0.127 & 0.505 / 0.147 & 0.479 / 0.108 & 0.466 / $\mathbf{0.103}$ \\
		\hline
	\end{tabular}
	\label{performance}
\end{table*}

The statistical results on public datasets are summarized in Table.\ref{performance}. CT-VLO (ours) denotes the LiDAR-only mode of the proposed algorithm, while CT-VLIO (ours) represents the IMU-LiDAR fusion mode. For the LO mode, the covariance of the continuous trajectory fitting error is set using fixed hyperparameters. 
%Although parameters vary slightly across datasets, their magnitudes are maintained at approximately $10^{-5}$ for rotation covariance and $10^{-4}$ for translation covariance.
For the LIO mode, the covariance of the continuous trajectory fitting error is computed using  eq.(\ref{fitting_error_rotation}).

According to the results presented in the table, the proposed method achieves the best performance in terms of APE metrics across multiple sequences from different public datasets. Among the compared methods in the MCD dataset, the MAX metric is not reported, as their results are sourced from \cite{SLICT2, RESPLE}, which do not include the MAX metric. Therefore, to ensure a fair comparison, only the RMSE metric IS compared here, while the MAX metric of the proposed method are listed solely for illustration.

\subsubsection{\textbf{MCD dataset}}
Among these, the MCD dataset comprises campus scenes, and the proposed method attains the best performance on most sequences within this dataset. For the `ntu-night13` sequence, due to the time delay between the IMU and LiDAR measurements, the continuous-time-based method, which is sensitive to the timestamps of measurements, experiences a degradation in system estimation performance after incorporating IMU measurements. 

\subsubsection{\textbf{M2UD dataset}} 
For the M2UD dataset, several challenging sequences were selected. Among them, the `aggressive04` sequence was collected using an all-terrain vehicle traversing rough terrain, including scenarios such as going up and down stairs and undergoing large-scale pitch motion. In this sequence, only R-LIO and the proposed algorithm were able to complete the entire sequence, with the proposed algorithm achieving better performance. Notably, CT-VLO outperformed CT-VLIO in this sequence, primarily because the IMU measurements exhibited significant abrupt changes during rough terrain traversal, and since no specific processing was applied to handle such IMU data, the fusion of IMU measurements actually degraded the overall performance. The `plaza03` sequence involves multiple floors within a shopping mall. Since all methods failed after entering the elevator at 1350 seconds, the time range for this sequence was fixed to $[0, 1350]$. Among the methods, F-LIO, while not achieving the best performance, demonstrated relatively stable results. However, CLIC and F-LIVO2 were unable to complete this sequence, and R-LIO exhibited instability, failing repeatedly. The reported results for R-LIO in this sequence are the best obtained from a limited number of successful runs.

\subsubsection{\textbf{MARS-LVIG dataset}}
For the MARS-LVIG dataset, three sequences involving medium-speed ($8$ m/s) to high-speed ($12$ m/s) motion (including GNSS information) were selected. 
In the `HKairport02` sequence, during takeoff or landing, LiDAR measurements are severely degraded, and all methods exhibit certain drift. After incorporating IMU measurements, smaller drift can be achieved, improving the robustness of the system.
In the `AMtown03` sequence, which includes flat-area takeoff and landing as well as rapid large-scale turns, LiDAR degeneration occurs. In such scenarios, the CLIC method exhibits significant drift and fails to complete the sequence.

\subsubsection{\textbf{Diter++ dataset}}
For the Diter++ dataset, a total of two sequences are used. The `Forest-new` sequence involves short-distance operation within a forest area with relatively complex terrain, and all selected methods achieved good performance. The `Park-in-day-cam-free` sequence involves navigation through a combination of forest areas, buildings, and long corridors, with large-scale and long-duration operation (2160 seconds). Among the methods, CLIC could only maintain operation for a limited period and failed to complete the sequence. In contrast, CT-VLO successfully completed the entire sequence, achieving the best APE metric. The complete operation results for the sequence are presented in the supplementary video material.

\begin{figure*}[htbp]
	\centering
		\subfloat[CLIC]{
				\includegraphics[width=0.315\textwidth]{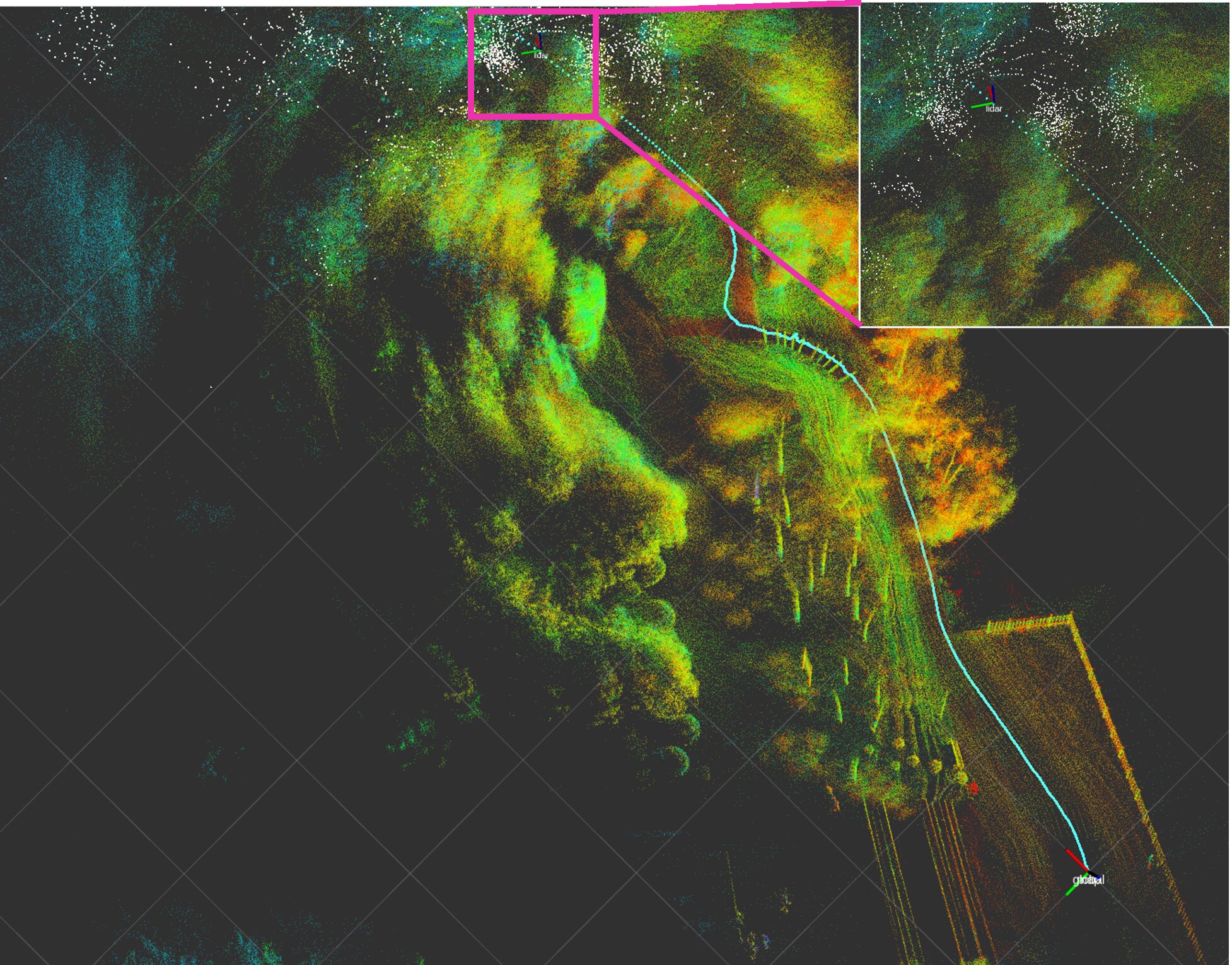}
			}
	%	\hfill
		\subfloat[F-LIO2]{
				\includegraphics[width=0.315\textwidth]{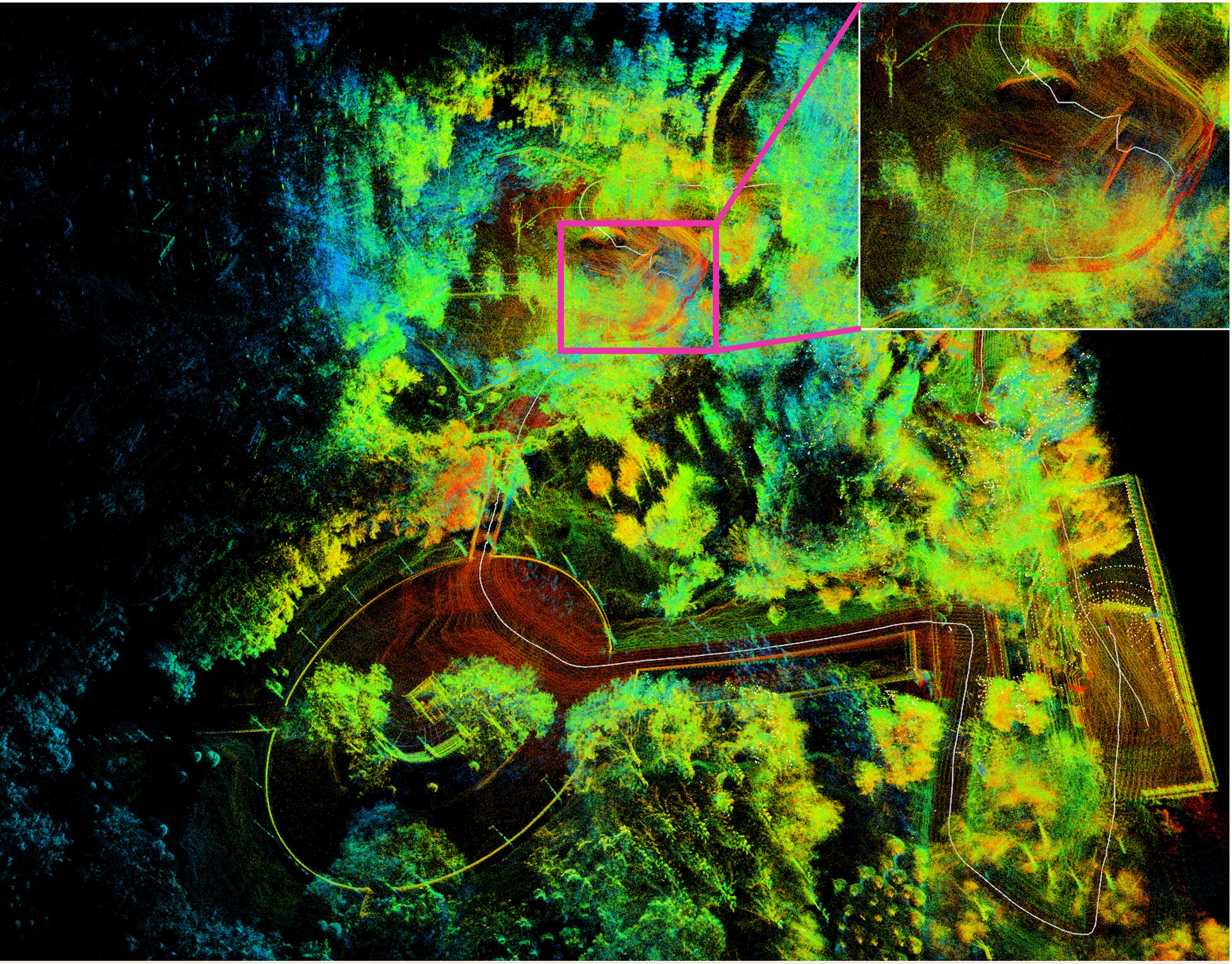}
			}
	%	\hfill
		\subfloat[F-LIVO2]{
				\includegraphics[width=0.315\textwidth]{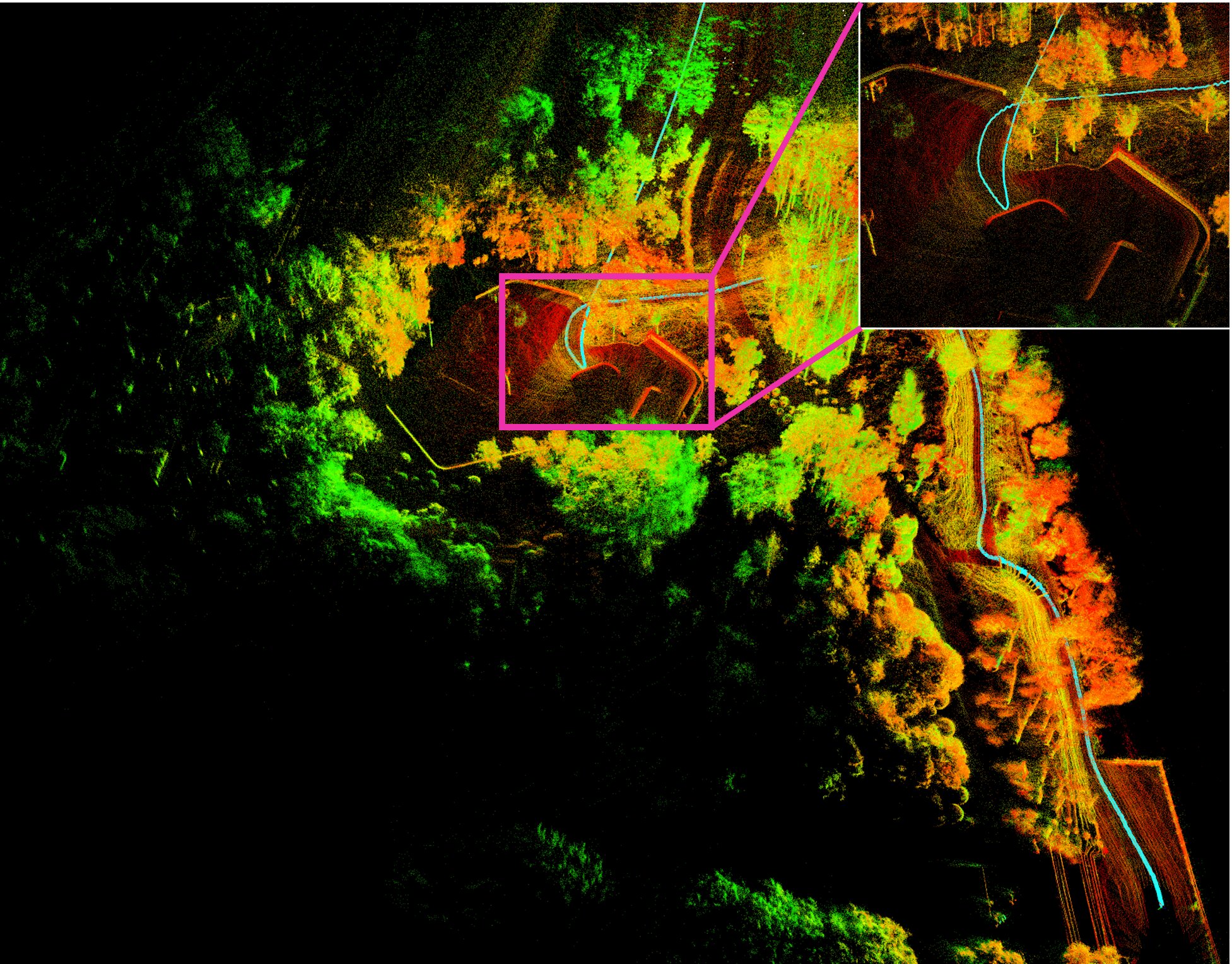}
			}
		\hfill
		\subfloat[RESPLE]{
				\includegraphics[width=0.315\textwidth]{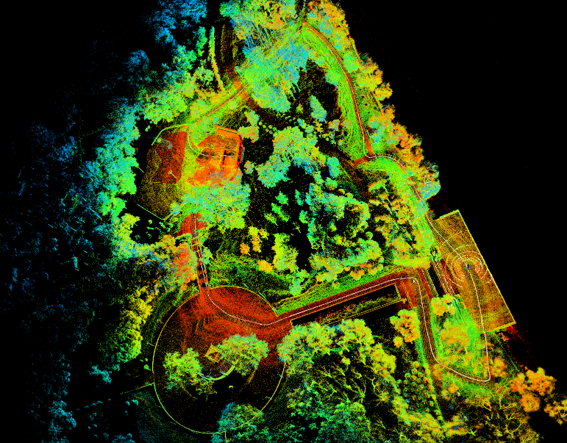}
			}
	%	\hfill
		\subfloat[CT-VLO]{
				\includegraphics[width=0.315\textwidth]{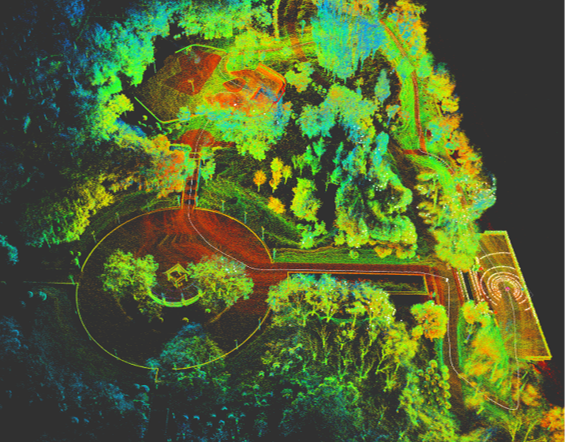}
			}
	%	\hfill
		\subfloat[CT-VLIO]{
				\includegraphics[width=0.315\textwidth]{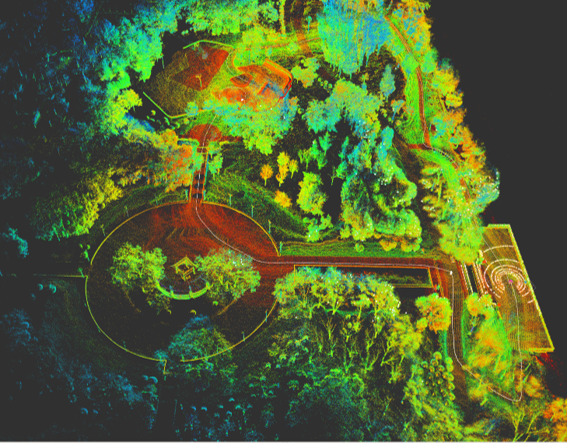}
			}
		\hfill
	\subfloat[CLIC]{
		\includegraphics[width=0.315\textwidth]{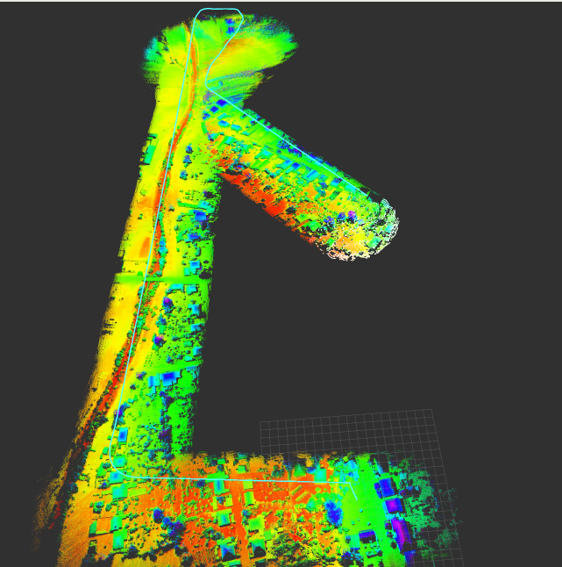}
	}
	%	\hfill
	\subfloat[F-LIO2]{
		\includegraphics[width=0.315\textwidth]{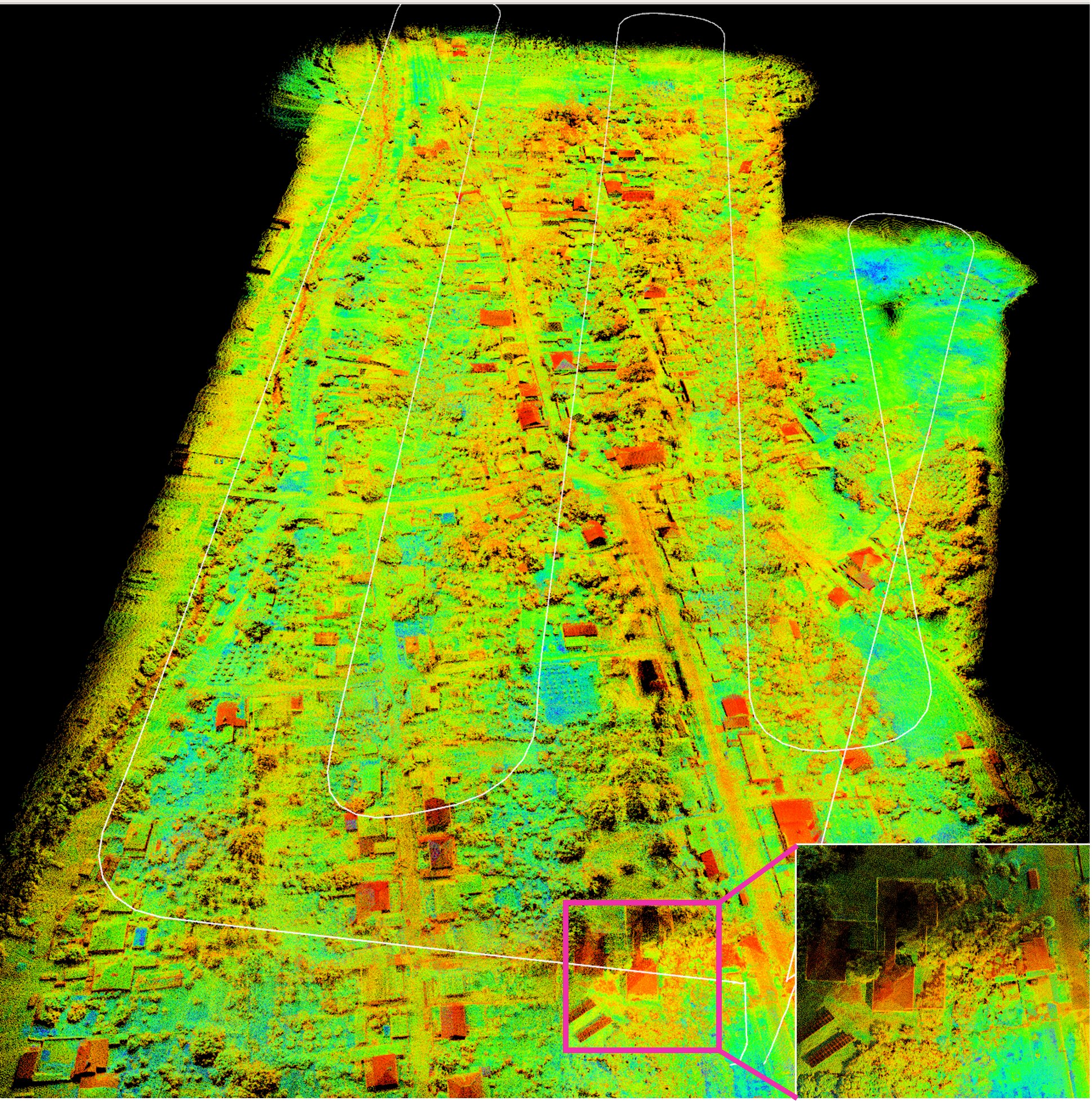}
	}
	%	\hfill
	\subfloat[F-LIVO2]{
		\includegraphics[width=0.315\textwidth]{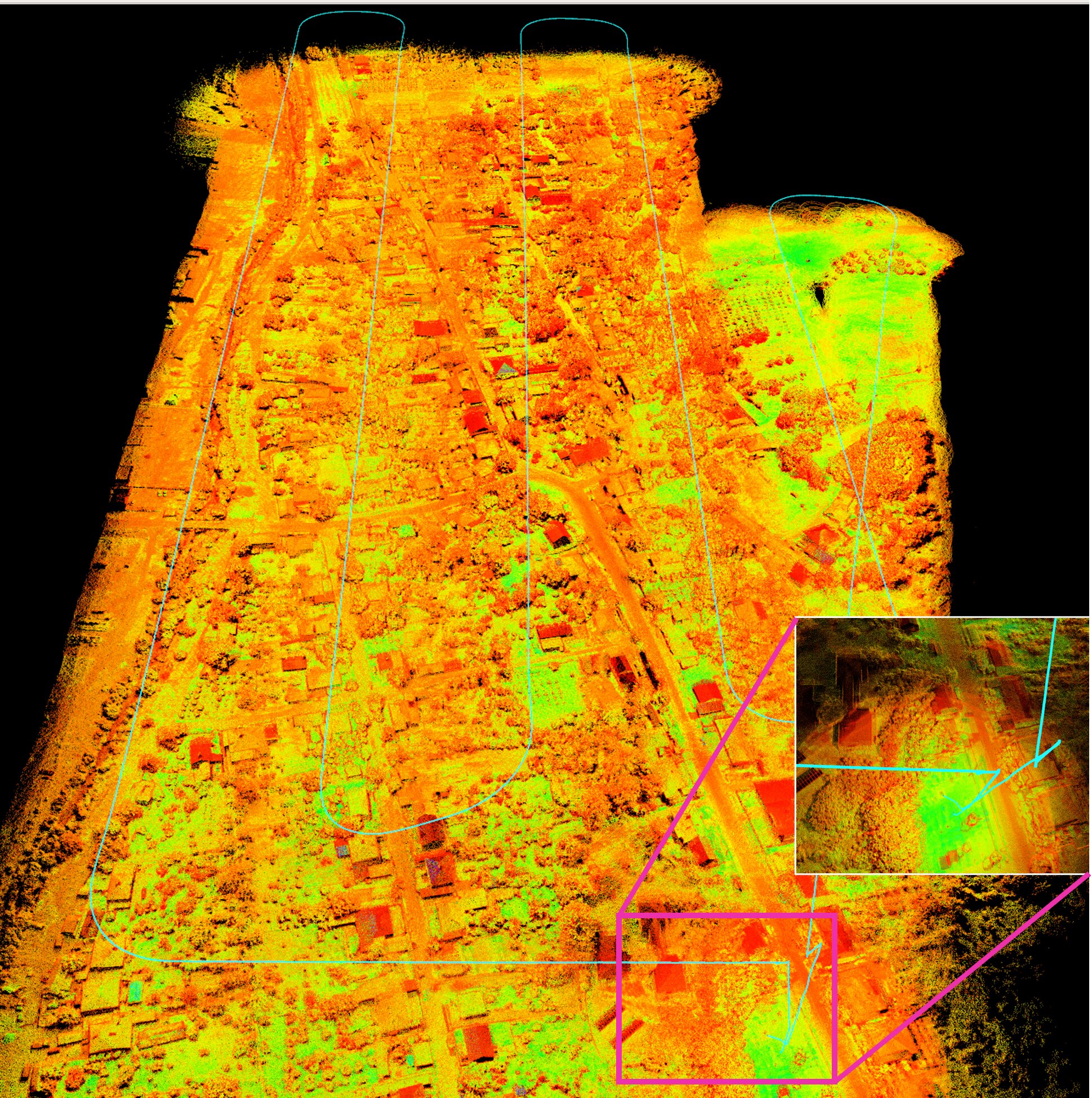}
	}
	\hfill
	\subfloat[RESPLE]{
		\includegraphics[width=0.315\textwidth]{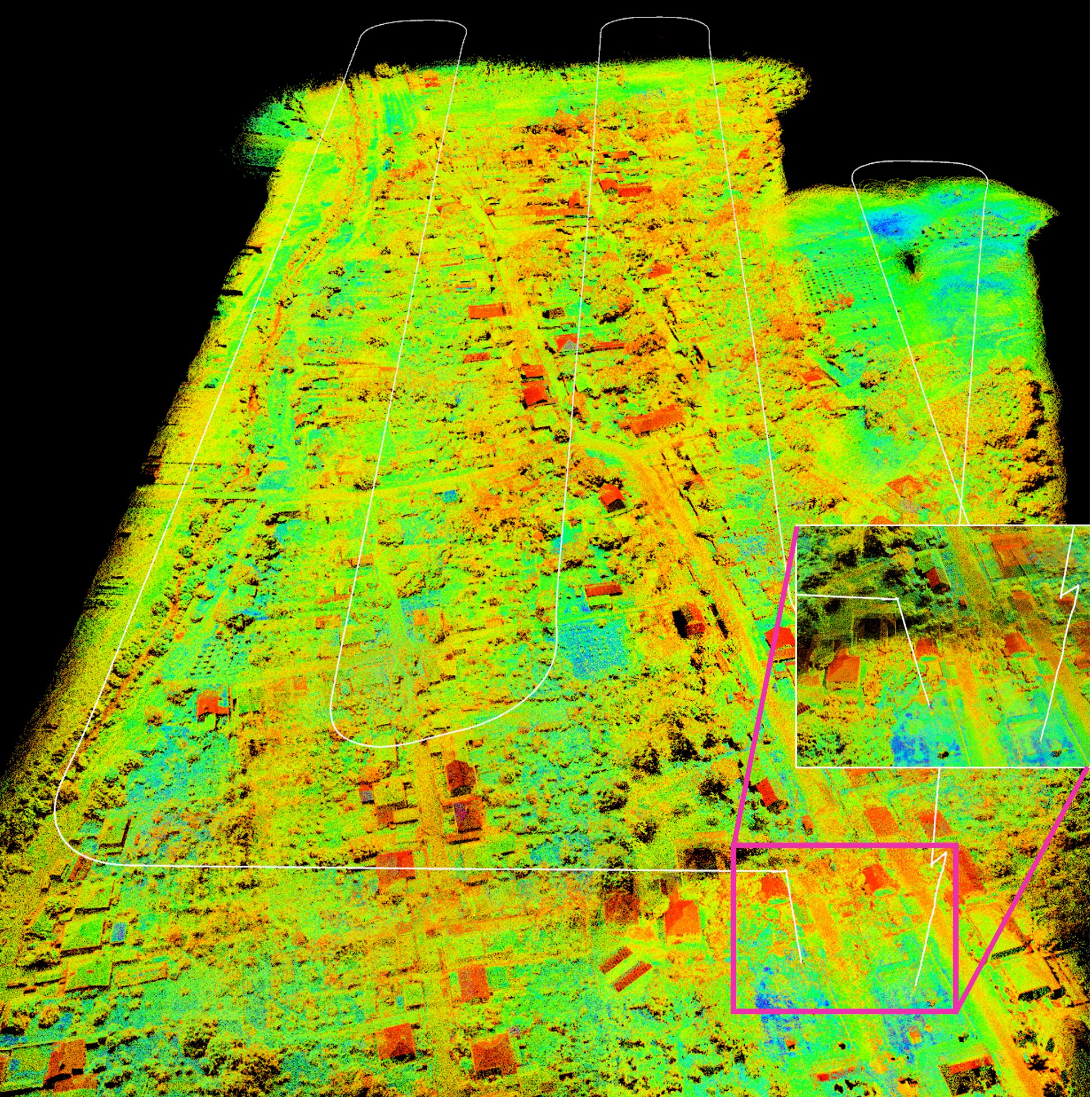}
	}
	%	\hfill
	\subfloat[CT-VLO]{
		\includegraphics[width=0.315\textwidth]{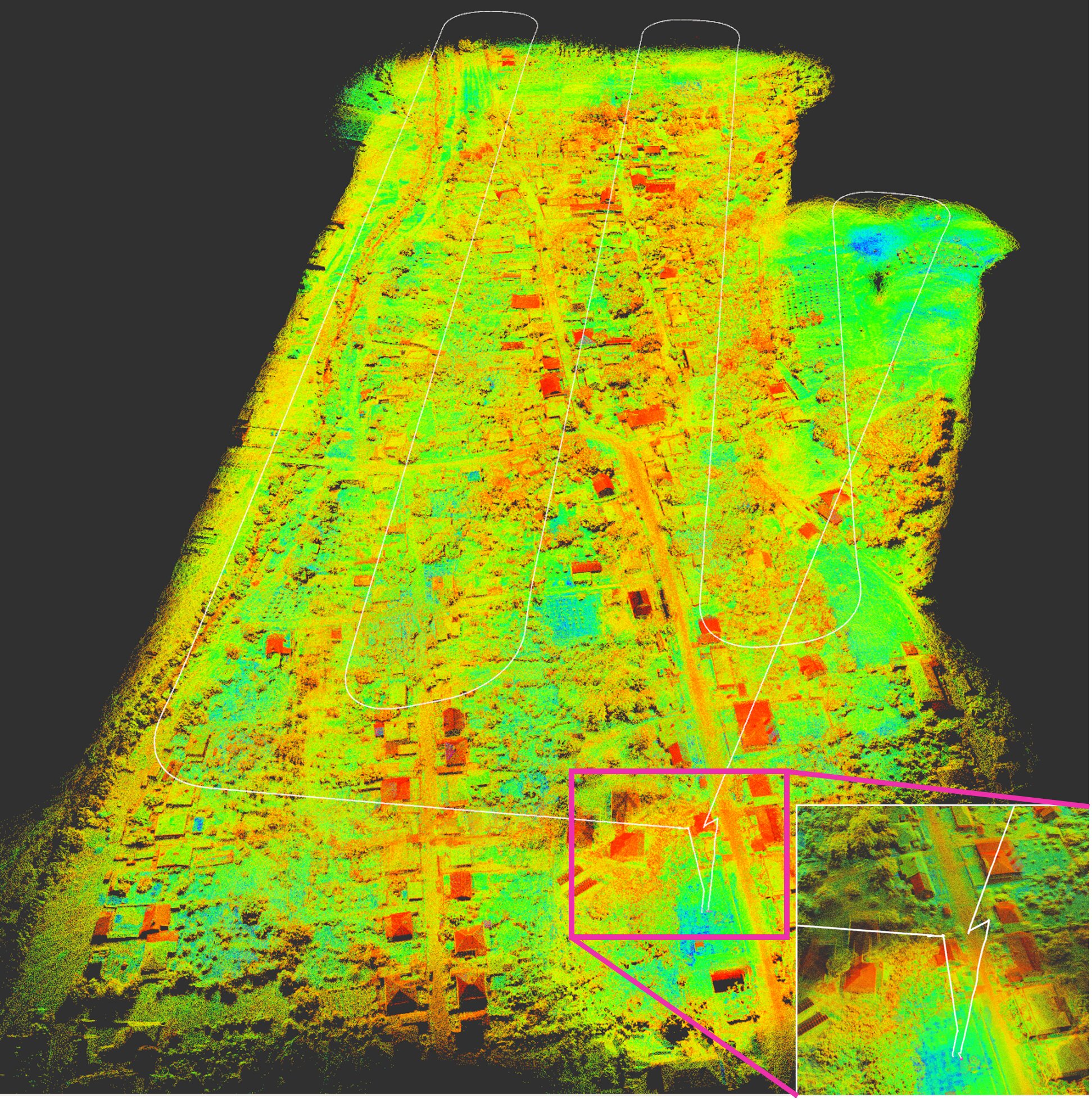}
	}
	%	\hfill
	\subfloat[CT-VLIO]{
		\includegraphics[width=0.315\textwidth]{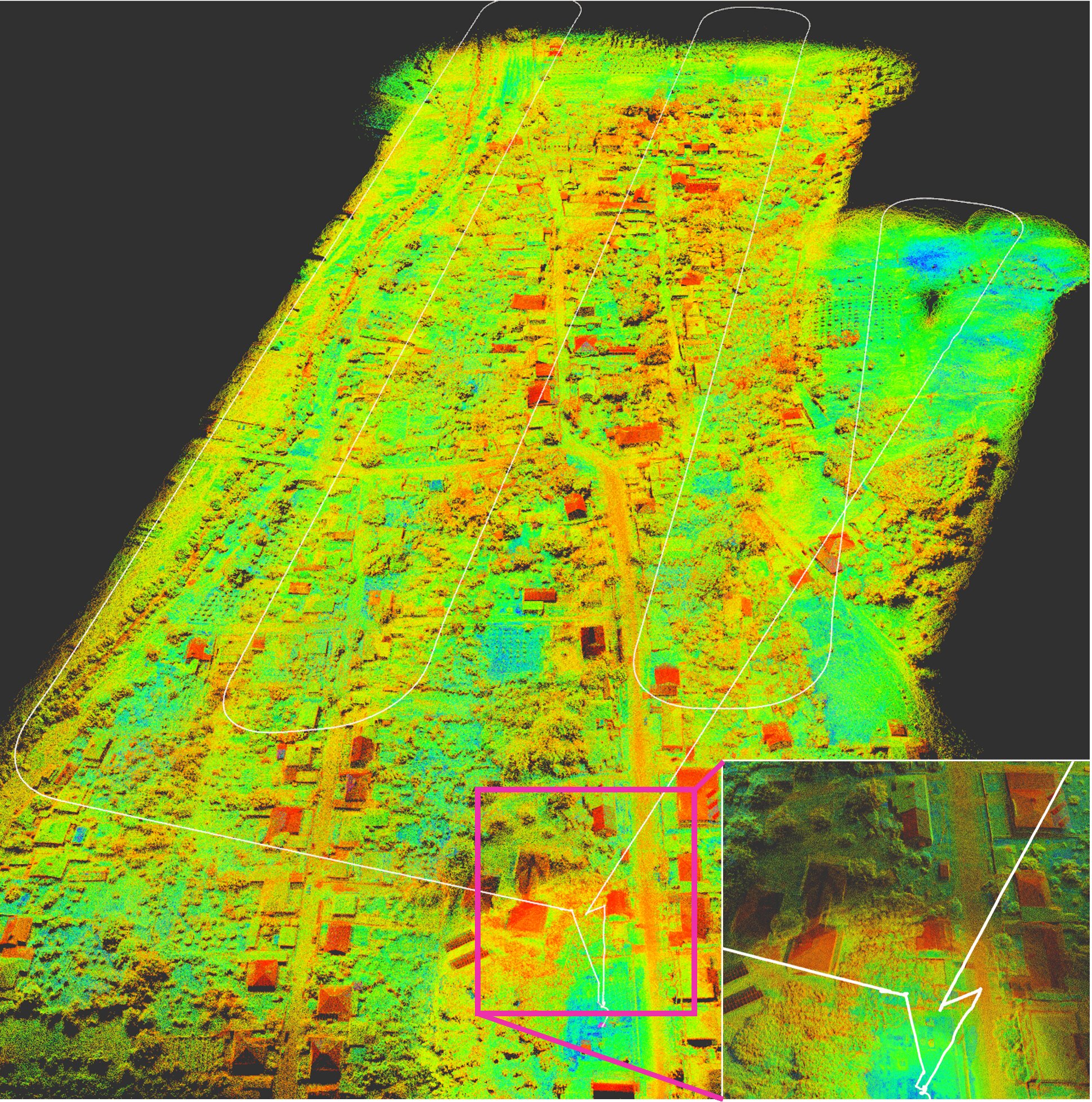}
	}
	\caption{ Performing results comparison of different methods on the `aggressive04` and `AMtown03` sequence.}
	\label{analysis}
\end{figure*}

To provide a more intuitive illustration of algorithm performance, the challenging sequence `aggresive04` and `AMtown03` are selected for further analysis. The mapping results on these sequences are presented in Figure.~\ref{analysis}.
On the `aggressive04` sequence, the CLIC method begins to exhibit pose estimation oscillations after traversing the first staircase and eventually fails completely. Both F-LIO2 and F-LIVO2 struggle to maintain consistent pose estimation when encountering severe pitch variations, resulting in abrupt attitude jumps. In contrast, RESPLE, CT-VLO (ours), and CT-VLIO (ours) all sustain stable pose estimation throughout the sequence.
On the `AMtown03` sequence, CLIC loses track during the first rapid large-angle turn. The remaining methods successfully complete the entire sequence. However, since none of these approaches incorporate loop closure techniques, all exhibit accumulated pose estimation errors. Comparatively, CT-VLO and CT-VLIO maintain relatively smaller cumulative errors, yielding clearer mapping results.

\subsection{Ablation study}

Based on the APE metrics, the results on public datasets demonstrate the performance of CT-VLO and CT-VLIO. To gain deeper and more direct insights into the impact of each proposed module, ablation studies are conducted on three sequences with different LiDAR sensors: MCD: ntu-day01 (Mid-70); M2UD: aggressive04 (Velodyne-16); and MARS-LVIG: AMtown03 (Avia). The CT-VLO method (LiDAR-only mode) is employed in these ablation experiments to eliminate the influence of IMU measurements. The effects of fitting error modeling, voxel features, and the re-estimation policy are investigated. When the re-estimation policy is enabled, the maximum number of re-estimations is set to 5 for all cases. Ten independent runs are performed on each sequence, with results summarized in Table.\ref{ablation}. The reported APE-related metrics include maximum deviation, mean, and RMSE, all in meters (m). Average processing time per frame is reported in milliseconds (ms). Success rate (S-rate) is computed as $SC / TC$, where $SC$ denotes success count and $TC$ denotes total count. Ablation configurations are denoted as follows: wo $fit-err$ (without cubic spline fitting error consideration); wo $re$ (without re-estimation policy); wo $voxel$ (without voxel features); wo $re \& voxel$ (without both re-estimation policy and voxel features); and w $all$ (full CT-VLO method). The APE metrics reported correspond to the best results among the ten runs.

\begin{table}[htbp]
	\centering
	\caption{Ablation Experiments of CT-VLO on datasets}
	\renewcommand{\arraystretch}{1.2} 
	\begin{tabular}{M{1.8cm} | M{2.5cm} M{1.4cm} M{1.2cm}}
		\hline
		\textbf{MCD} & \textbf{Max/Mean/RMSE} & \textbf{Ave-time} & \textbf{S-rate}  \\
		\hline
		wo fit-err & - & - & 0/10 \\
		wo re & 2.278/1.072/1.168 & 27.5316 & 6/10 \\
		wo voxel & 1.779/0.826/0.906 & 27.2216 & 10/10 \\
		wo re\&voxel & 2.504/0.885/1.009 & 27.4204 & 4/10 \\
		w all & 2.635/0.878/0.994 & 27.3120 & 9/10 \\
		\hline
		\textbf{M2UD} & \textbf{Max/Mean/RMSE} & \textbf{Ave-time} & \textbf{S-rate} \\
		\hline
		wo fit-err & - & - & 0/10  \\
		wo re & 2.046/0.601/0.691 & 33.796 & 8/10 \\
		wo voxel &2.081/0.593/0.686 & 33.497 & 10/10 \\
		wo re\&voxel & 2.097/0.598/0.688 & 33.931 & 10/10 \\
		w all & 2.072/0.586/0.677 & 33.565 & 10/10 \\
		\hline
		\textbf{MARS-LVIG} & \textbf{Max/Mean/RMSE} & \textbf{Ave-time} & \textbf{S-rate} \\
		\hline
		wo fit-err & - & - & 0/10 \\
		wo re & - & - & 0/10 \\
		wo voxel & 18.959/2.704/2.939 & 62.2079 & 10/10 \\
		wo re\&voxel & - & - & 0/10 \\
		w all & 3.972/2.204/2.299 & 62.498 & 10/10 \\
		\hline
	\end{tabular}
	\label{ablation}
\end{table}

\begin{figure}[htbp]
	\centering
	\includegraphics[width=0.5\textwidth]{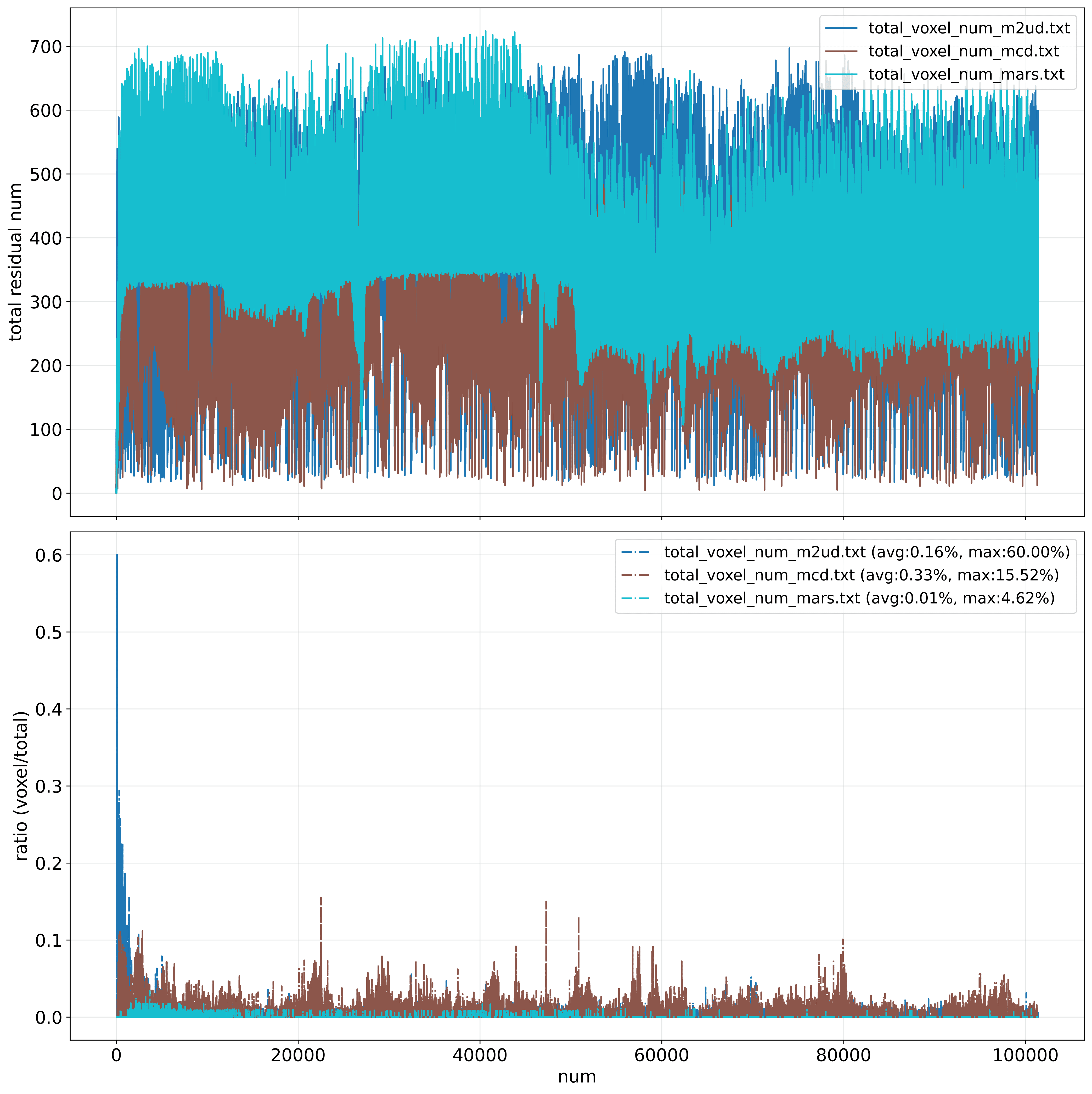}
	\caption{Statistical results of total and  voxel residual numbers.}
	\label{residual_num}
\end{figure}

\subsubsection{\textbf{Fitting Error}}
In the algorithm design, the error introduced by cubic B-spline trajectory fitting is explicitly considered. According to eq.(\ref{point_to_plane}), this error propagates into the observation noise, thereby increasing the uncertainty of the observations. The results in Table.\ref{ablation} indicate that under the current framework, the system fails to operate properly when the continuous trajectory fitting error is not accounted for.

\subsubsection{\textbf{Voxel Feature}}
As shown in Table.\ref{ablation}, the incorporation of voxel features increases the information available to the system. However, the results in MCD indicate that on this dataset, voxel features lead to a slight decrease in both system stability and performance. From the results in M2UD, it can be observed that when the re-estimation policy is not applied, using only voxel features marginally reduces system stability; however, when both the re-estimation policy and voxel features are employed together, system performance improves. According to the results in MARS-LVIG, the introduction of voxel features further enhances estimation performance. 

Figure.\ref{residual_num} illustrates the relationship between the total number of residuals and the number of residuals corresponding to voxel features during runtime. Combining the results from Table.\ref{ablation}, it can be concluded that when the proportion of voxel features is high, system stability and performance tend to degrade. Conversely, when the proportion of voxel features is relatively low, they contribute positively to improving estimation performance.

\subsubsection{\textbf{Re-estimation Policy}}
As shown in MCD, the re-estimation policy improves the system success rate regardless of whether voxel features are utilized. According to the results in M2UD, when combined with voxel features, the re-estimation policy further enhances system robustness. The results in MARS-LVIG indicates that the system fails to operate properly without the re-estimation policy. Therefore, it can be concluded that the re-estimation policy significantly improves system robustness. Additionally, the tabulated results show that average per-frame processing time increases when the policy is not applied, further demonstrating its effectiveness in improving computational efficiency. To further investigate the impact of the re-estimation policy, a detailed analysis of the maximum number of re-estimations is provided in the discussion section.

%\begin{figure}[t]
%	\centering
%	\subfloat[Statistical results of re-estimate times]{
%		\includegraphics[width=0.5\textwidth]{pictures/experiments/re-estimate_compress.png}
%		\label{re-estimate:sub1}
%	}
%	\hfill
%	\subfloat[Statistical results of re-estimate times on the interval of (10000, 10200).]{
%		\includegraphics[width=0.5\textwidth]{pictures/experiments/re-estimate_local_compress.png}
%		\label{re-estimate:sub2}
%	}
%	\caption{ Performing results comparison of CT-VLO with different re-estimate times on AMtown03 sequence}
%	\label{re-estimate}
%\end{figure}

\begin{figure}[t]
	\centering
	\includegraphics[width=0.5\textwidth]{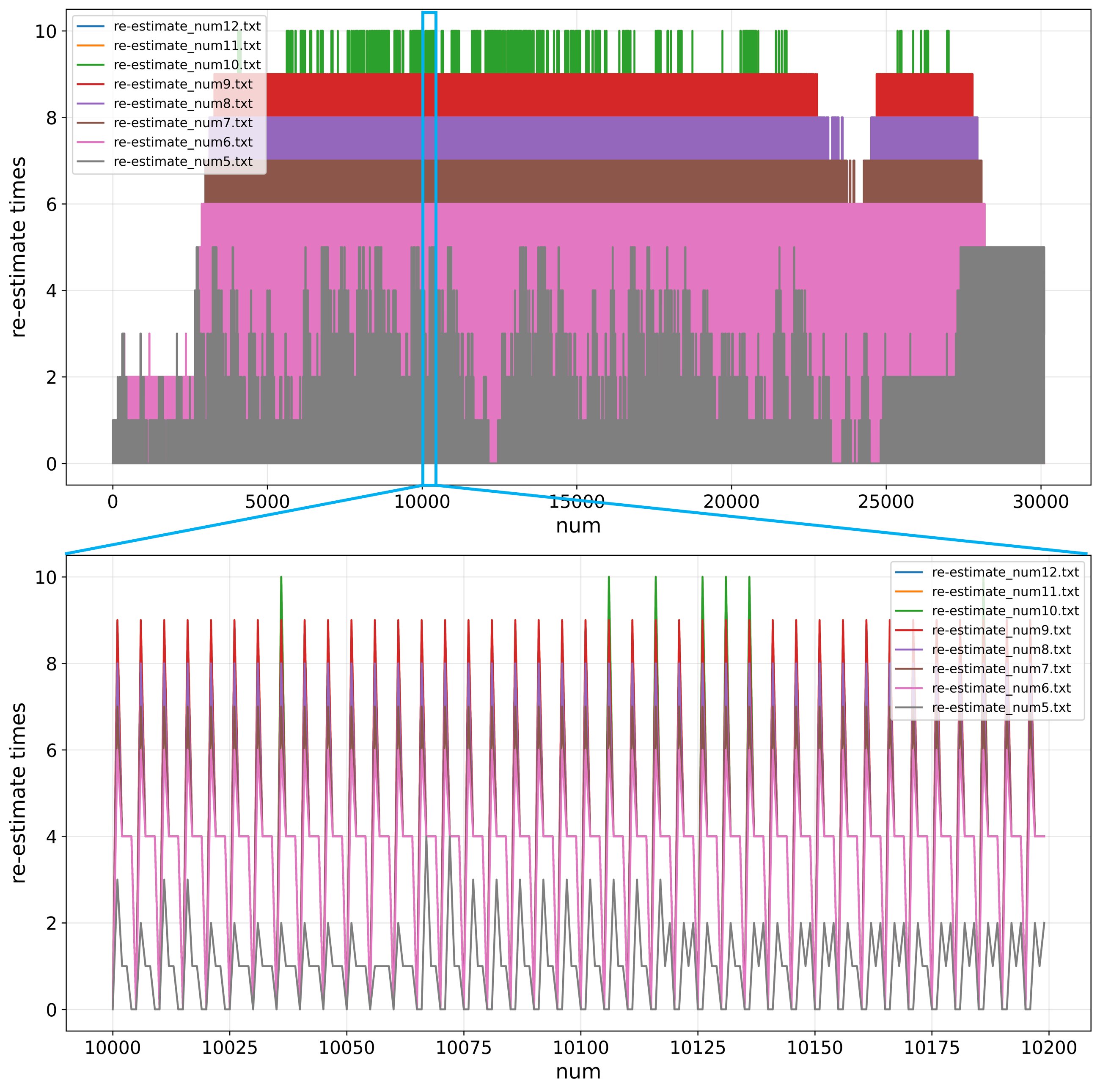}
	\caption{ Performing results comparison of CT-VLO with different re-estimate times on `AMtown03` sequence}
\label{re-estimate}
\end{figure}

\begin{table}[t]
	\centering
	\caption{Re-estimate policy of CT-VLO on MARS-LVIG dataset}
	\renewcommand{\arraystretch}{1.2} 
	\begin{tabular}{M{1.5cm} M{1.8cm} M{2.6cm} M{1.0cm}}
		\hline
		\textbf{AMtown03} &\textbf{re-estimate times} & \textbf{Max/Mean/RMSE} & \textbf{Ave-time} \\
		\hline
		CT-VLO & 1 & - & - \\
		CT-VLO & 2 & - & - \\
		CT-VLO & 3 & - & - \\
		CT-VLO & 4 & - & - \\
		CT-VLO & 5 & 3.971/2.204/2.299 & 62.4751 \\
		CT-VLO & 6 & 10.611/2.265/2.481 & 63.6368 \\
		CT-VLO & 7 & 29.316/3.186/3.989 & 66.2219 \\
		CT-VLO & 8 & 41.146/2.814/3.485 & 69.2019 \\
		CT-VLO & 9 & 7.416/3.459/3.721 & 71.3103 \\
		CT-VLO & 10 & $\mathbf{3.829/2.149/2.253}$ & 74.809 \\
		CT-VLO & 11 & 7.219/3.665/3.986 & 75.4734 \\
		CT-VLO & 12 & 7.219/3.665/3.986 & 75.0008 \\
		\hline
	\end{tabular}
	\label{re-estimate}
\end{table}

\begin{figure*}[ht]
	\centering
	\subfloat[Trajectory]{
		\includegraphics[width=0.31\textwidth]{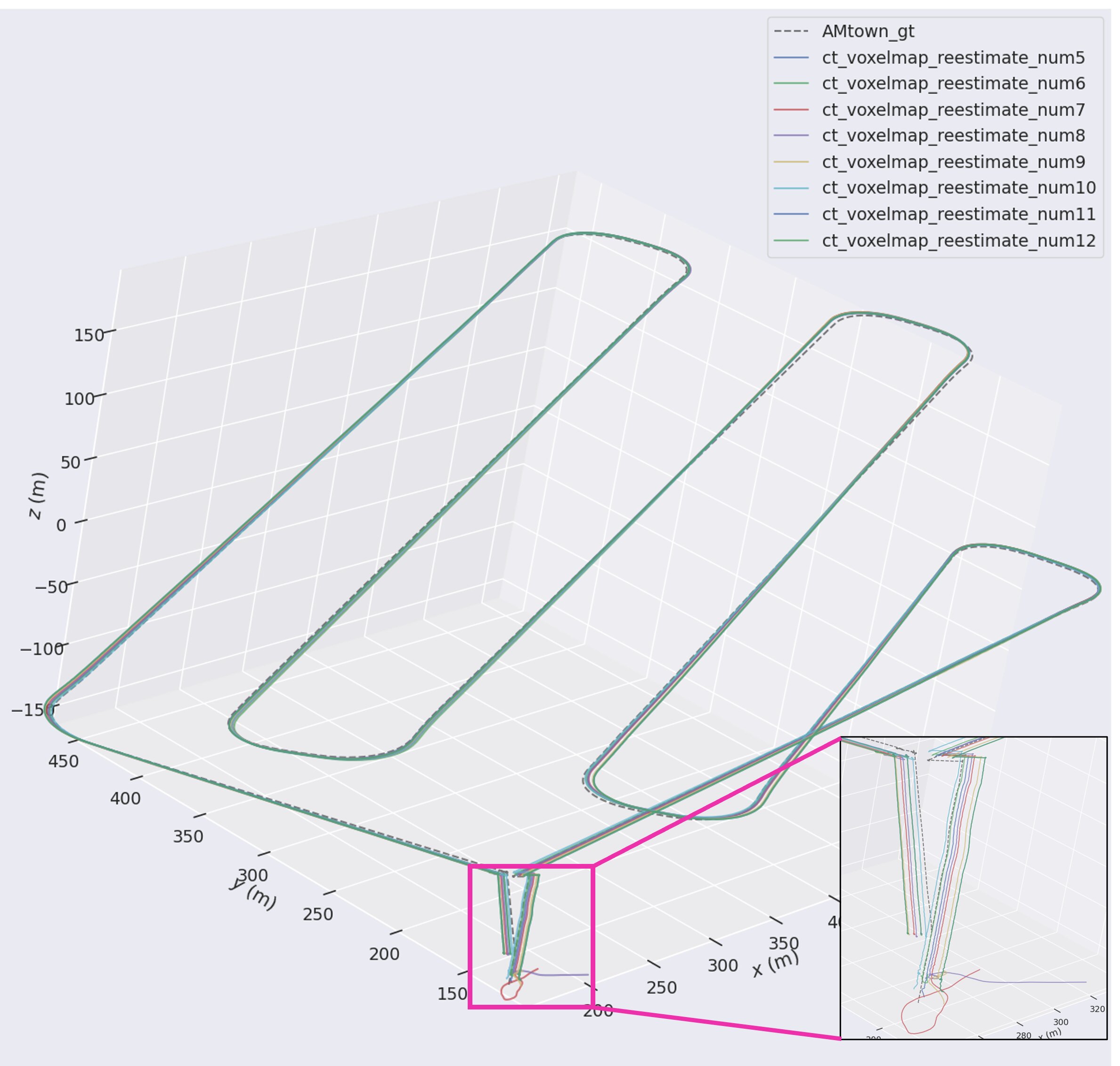}
	}
	\hfill
	\subfloat[Plot trajectories in the x, y, and z directions]{
		\includegraphics[width=0.31\textwidth]{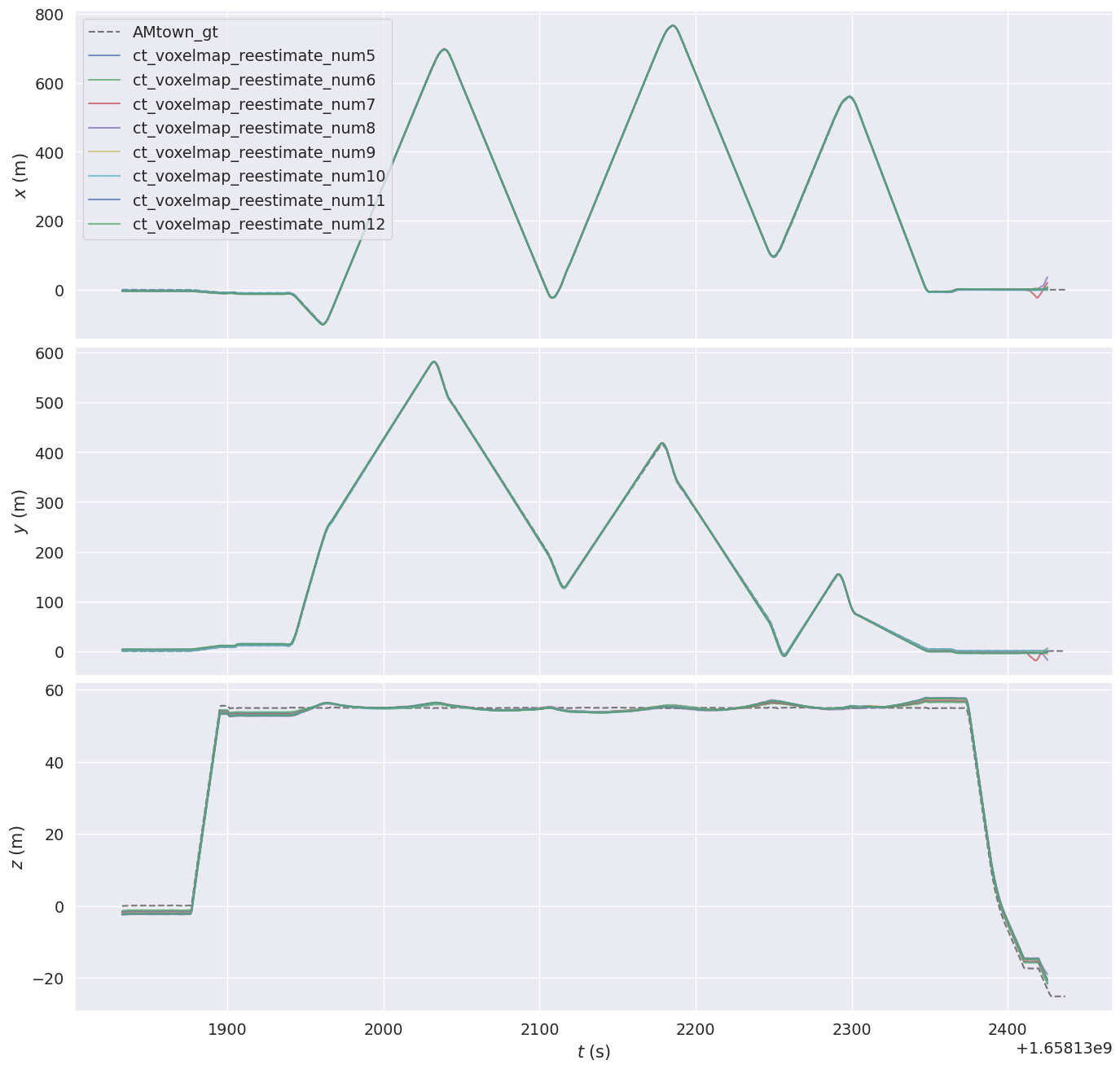}
	}
	\hfill
	\subfloat[Speed curve]{
		\includegraphics[width=0.31\textwidth]{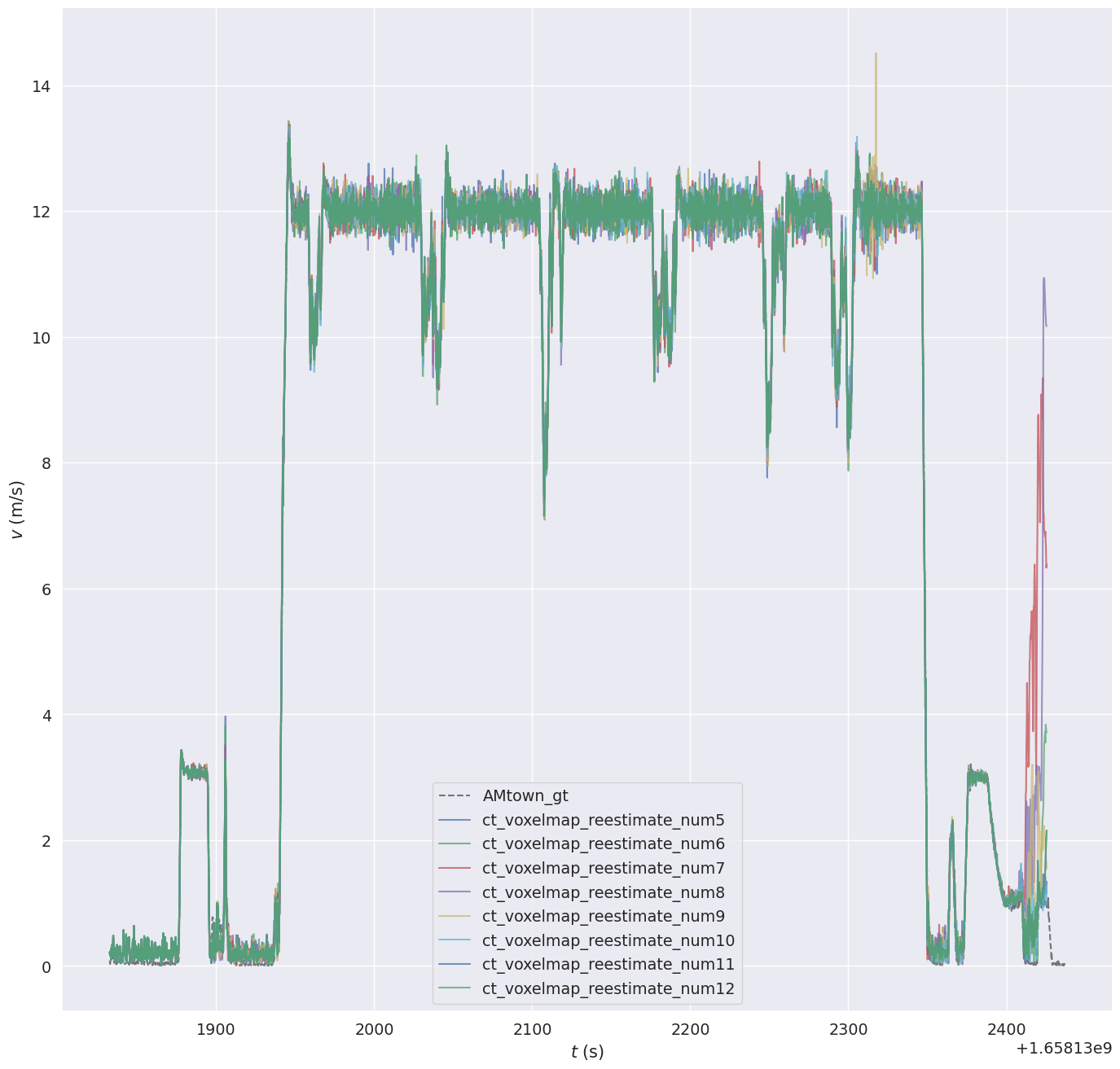}
	}
	\caption{ Performing results comparison of CT-VLO with different re-estimation times on `AMtown03` sequence.}
	\label{discussion}
\end{figure*}

\subsection{Discussion}

Through ablation studies, it is evident that the re-estimation policy significantly impacts system robustness. This section further discusses the effect of the maximum number of re-estimations within the re-estimation policy. 
%Since the AVIA LiDAR produces denser point clouds per scan, triggering the re-estimation policy more frequently.
The discussion is conducted on the `AMtown03` sequence, with results presented in Table.\ref{re-estimate}.

According to the results in Table.\ref{re-estimate}, when the maximum number of re-estimations is set too low, the majority of points in a scan are discarded, leading to system failure. As the maximum number of re-estimations gradually increases, both the number of utilized points per scan and the processing time increase accordingly. After reaching 10 re-estimations, the per-frame processing time stabilizes, with the best performance achieved at 10 re-estimations. When the maximum is set to 11 or 12, no significant change in processing time is observed, indicating that all points in the scan are being utilized. 

Figure.\ref{re-estimate} shows the actual number of re-estimations during sequence execution, where the horizontal axis represents the estimation index and the vertical axis represents the actual number of re-estimations. Under the current parameter settings, the actual maximum number of re-estimations is 10. The trajectory results are presented in Figure.\ref{discussion}, indicating that the impact of the re-estimation policy is amplified under LiDAR-degraded conditions, where a lower number of re-estimations contributes to system stability. Based on these findings, to balance performance and efficiency in practical applications, the maximum number of re-estimations is set to 5.

\subsection{wheel-legged experiments}

\begin{figure}[t]
	\centering
	\includegraphics[width=0.5\textwidth]{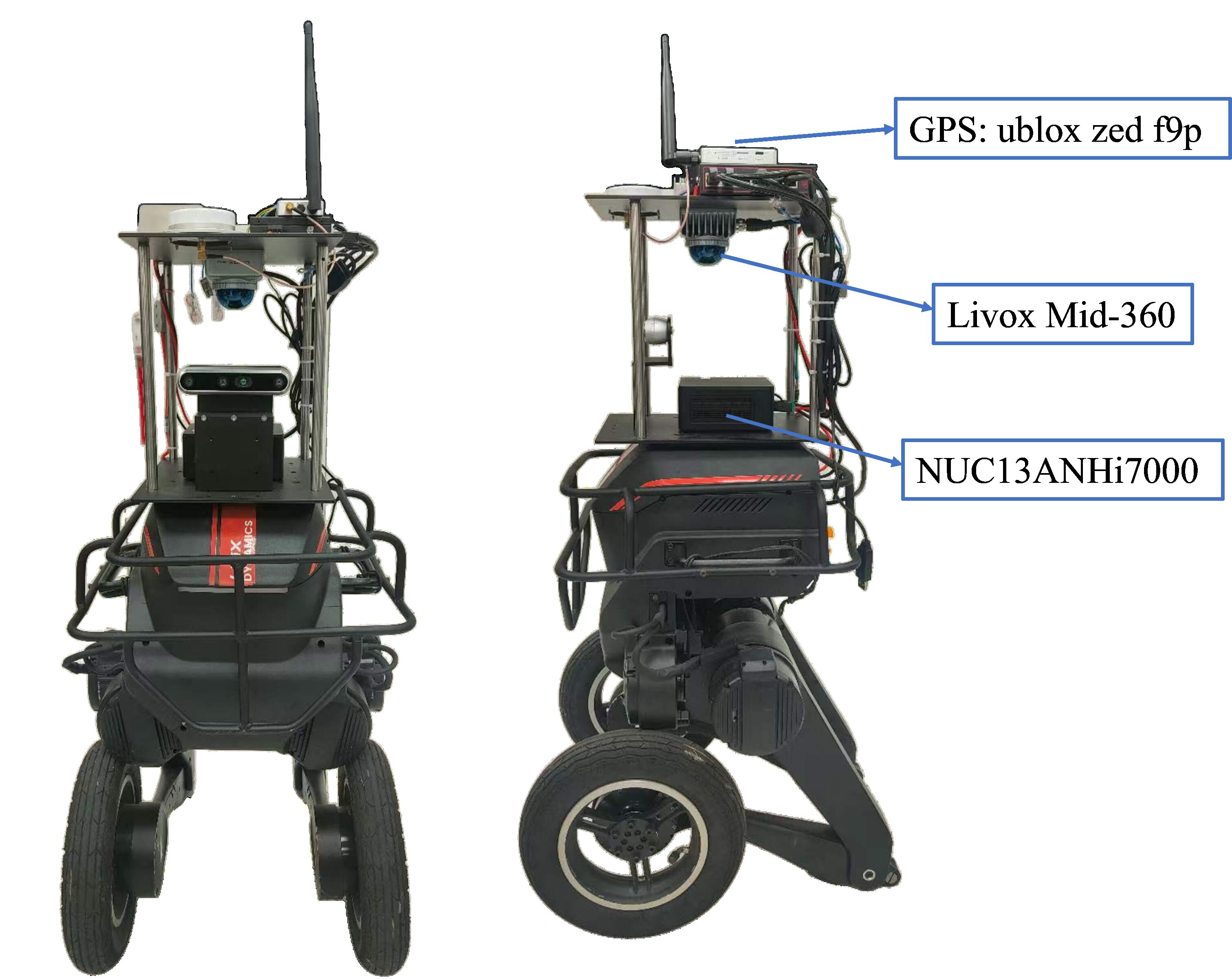}
	\caption{Description of the Experimental Wheel-Legged Robot Platform and Hardware. }
	\label{wheel-legged}
\end{figure}

To further demonstrate the algorithm's adaptability to different robot platforms (motion modes) and different LiDAR sensors, real-robot validation was conducted on campus using a wheel-legged robot. The robot platform is shown in Figure \ref{wheel-legged}.

\begin{figure*}[htbp]
\centering
\subfloat[Botanical01 env]{
	\includegraphics[width=0.31\textwidth]{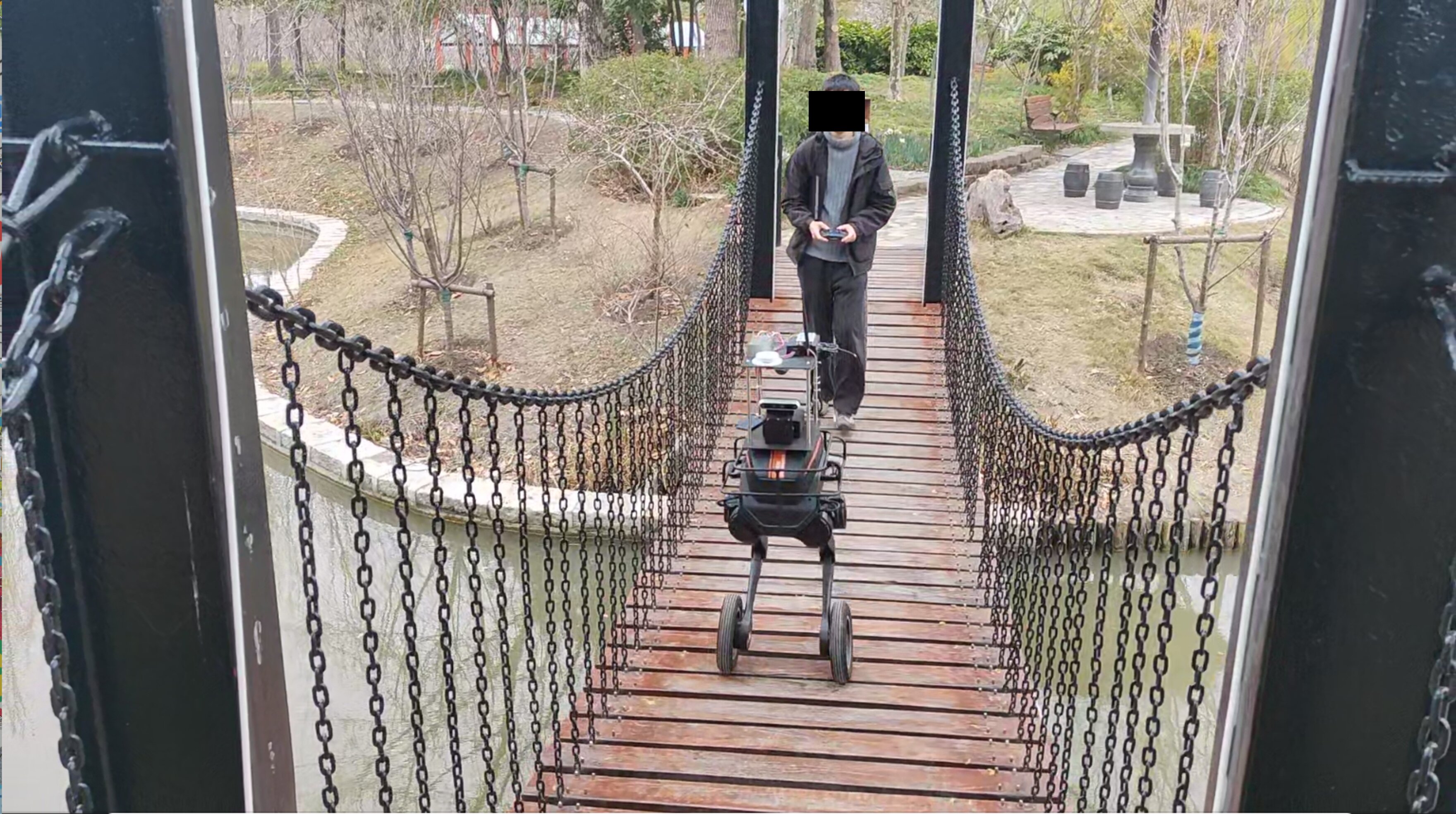}
}
\hfill
\subfloat[Botanical02 env]{
	\includegraphics[width=0.31\textwidth]{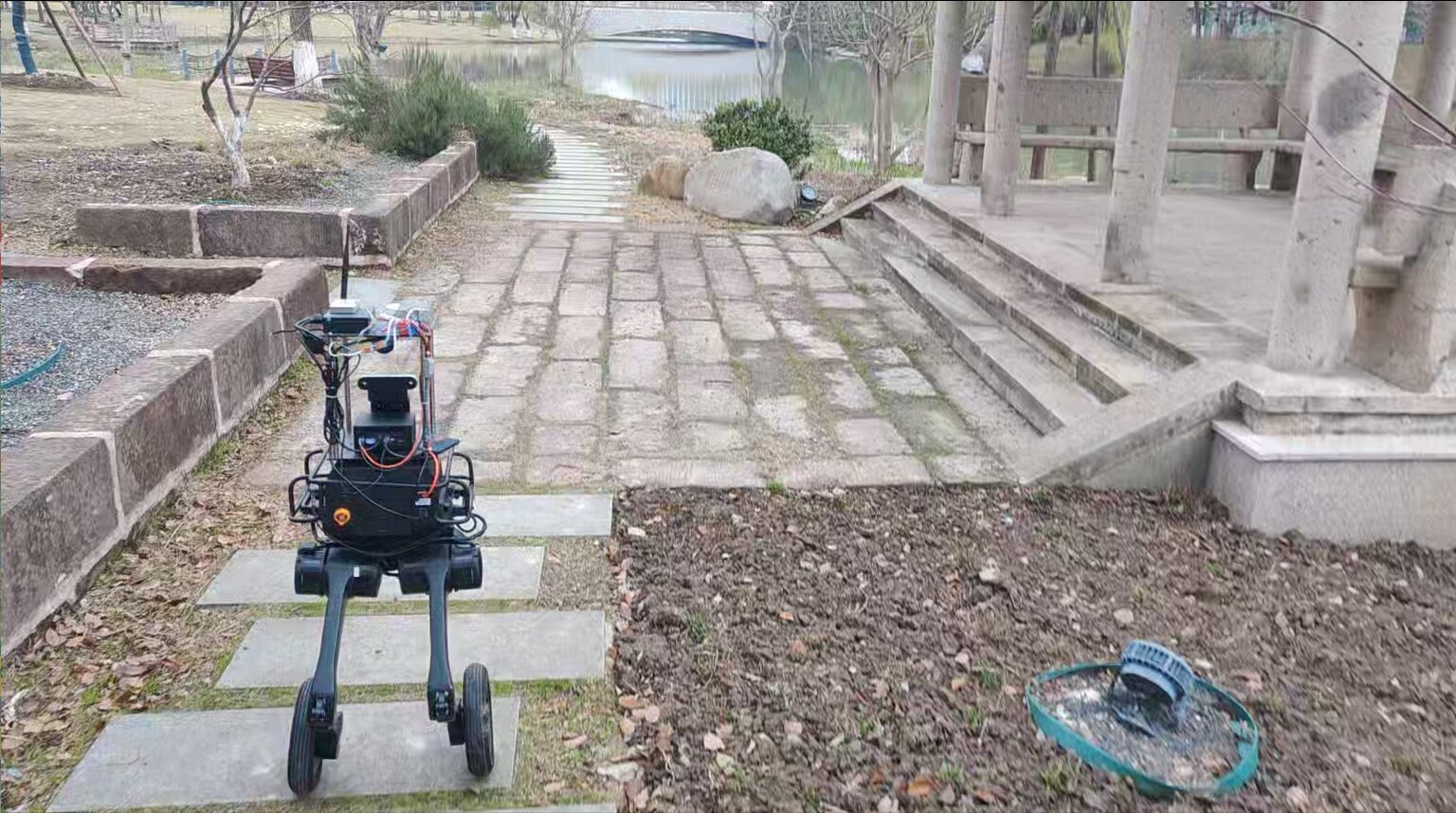}
}
\hfill
\subfloat[Botanical03 env]{
	\includegraphics[width=0.31\textwidth]{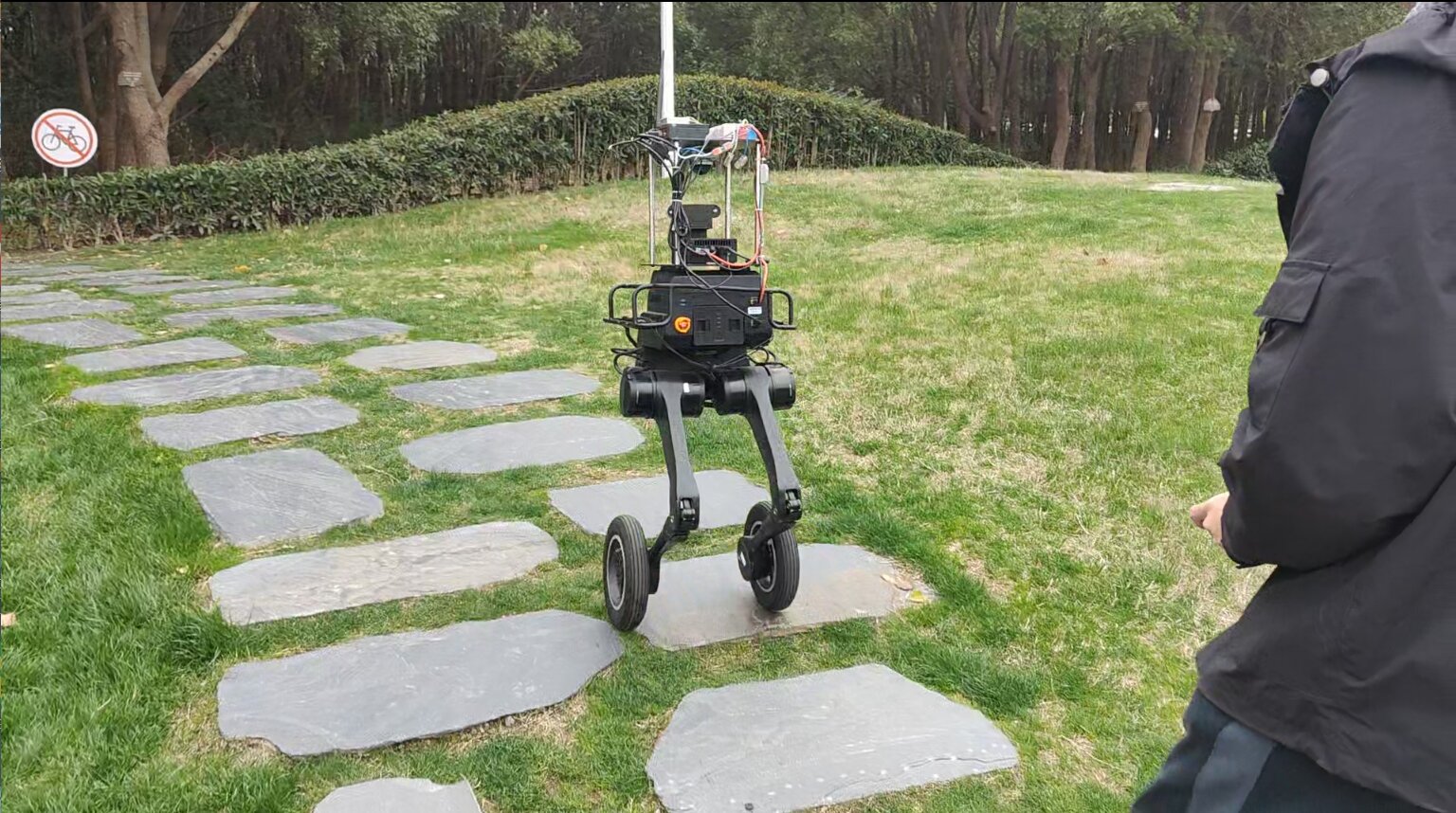}
}
\hfill
\subfloat[Botanical01 Trajectory]{
	\includegraphics[width=0.31\textwidth]{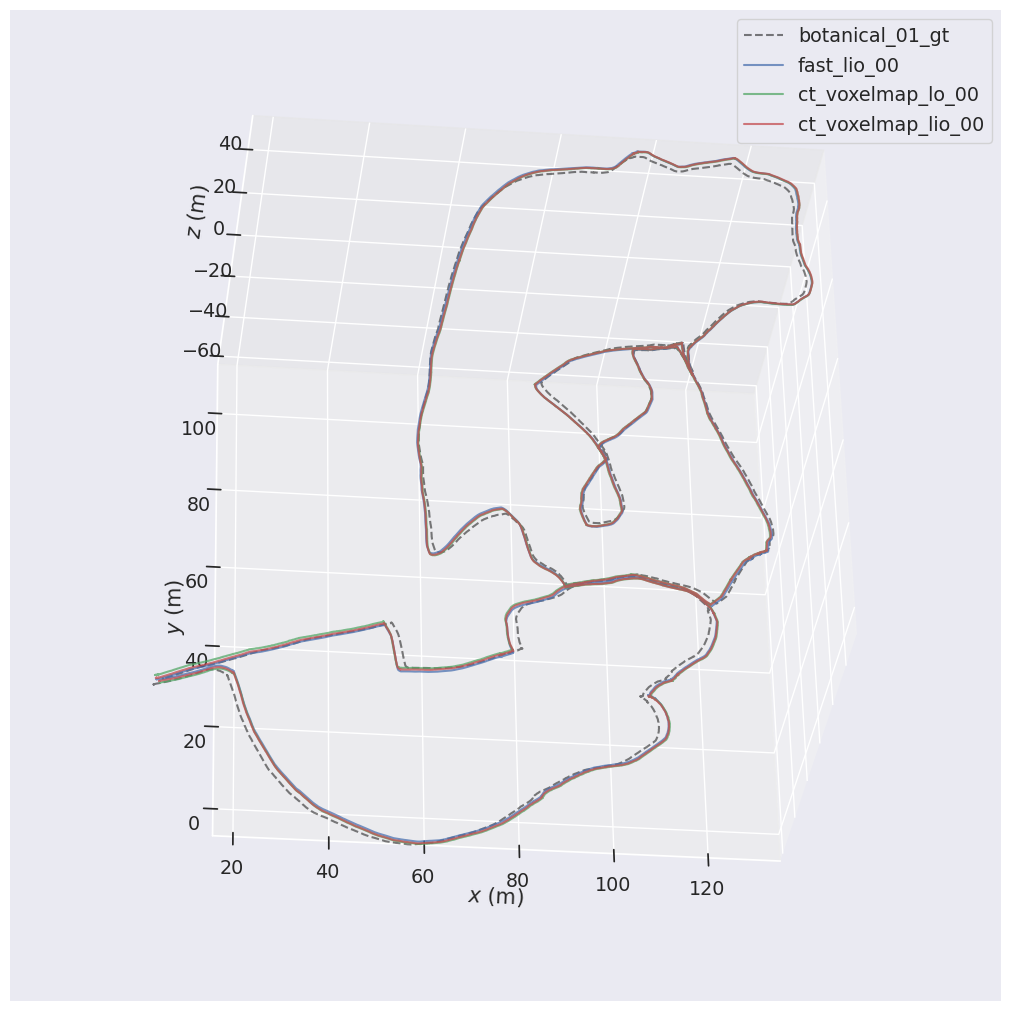}
}
\hfill
\subfloat[Botanical02 Trajectory]{
	\includegraphics[width=0.31\textwidth]{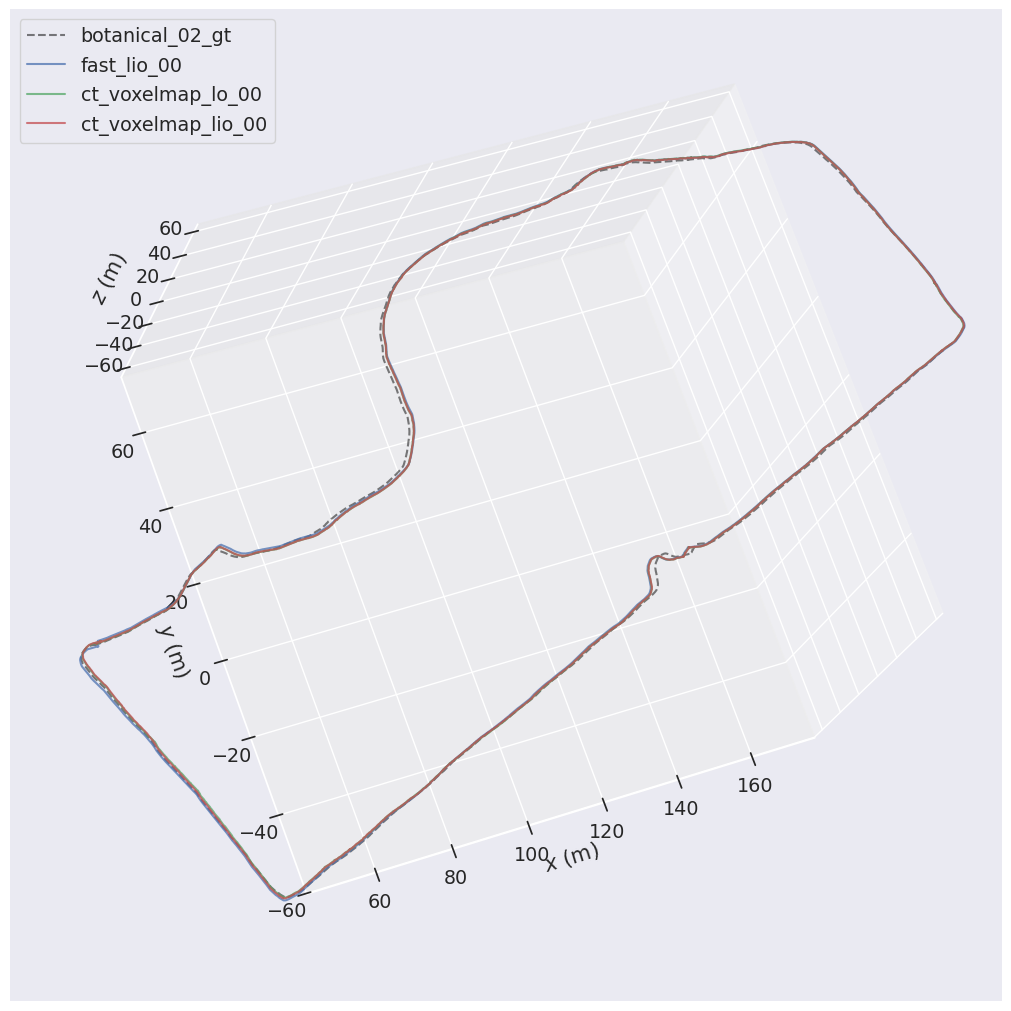}
}
\hfill
\subfloat[Botanical03 Trajectory]{
	\includegraphics[width=0.31\textwidth]{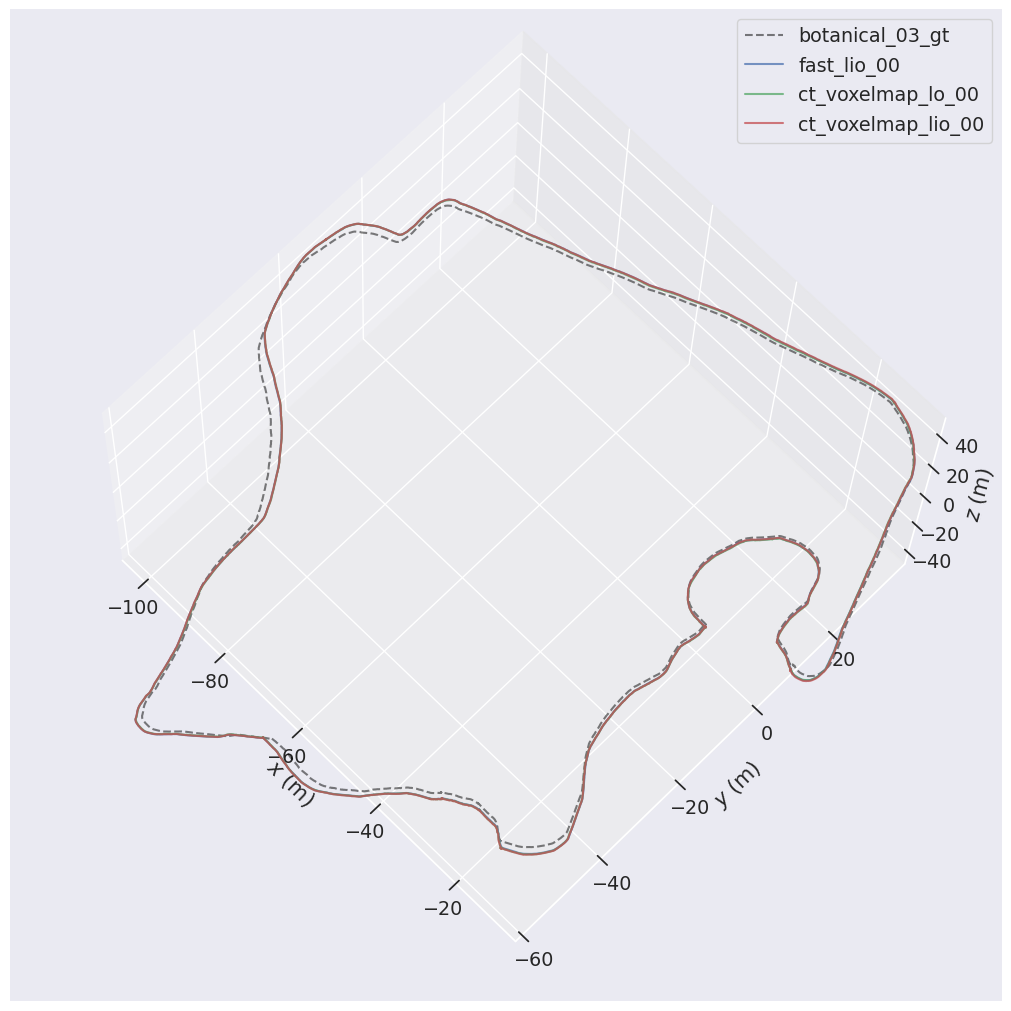}
}
\caption{Comparison of Trajectories Across Different Scenes in the Dataset.}
\label{botanical-env}
\end{figure*}

Real-world experiments were carried out on campus in challenging and bumpy terrain areas, including the botanical garden, riverside, and garden. Screenshots of the experimental environments and the experimental results are shown in Table \ref{botanical} and Figure \ref{botanical-env}, respectively.

Based on the real-robot experiments, better performance is still attainable on the wheel-legged robot with the Livox-Mid 360. 

\begin{table}[htbp]
	\centering
	\caption{APE Metric (RMSE, mster) on Our Self-Collected Botanical Dataset.}
	\renewcommand{\arraystretch}{1.4} 
	\begin{tabular}{c c c c}
		\hline
		\textbf{Dataset} &\textbf{F-LIO2} & \textbf{CT-VLO} & \textbf{CT-VLIO} \\
		\hline
		botanical01 & 1.853 & 1.723  & $\mathbf{1.711}$ \\
		botanical02 & 1.249 & 0.877  & $\mathbf{0.821}$ \\
		botanical03 & 2.093 & $\mathbf{2.008}$ & 2.025 \\
		\hline
	\end{tabular}
	\label{botanical}
\end{table}

\subsection{conclusion}

Through an in-depth analysis of the spline representation, this work improves upon the widely adopted cumulative-form B-spline on Lie groups, eliminating the need for additional boundary condition considerations (i.e., cases $j=0$ or $j=N-1$) when deriving Jacobians with respect to state variables. By estimating control point increments rather than directly estimating the control points themselves, the Jacobian computation is significantly simplified. Furthermore, by accounting for continuous trajectory fitting errors—either through online IMU-based estimation or predefined hyperparameters—the proposed system achieves stable operation. The introduction of voxel features within the voxel map enables the system to exploit additional information, thereby enhancing estimation performance. Finally, the proposed re-estimation policy, which involves multiple sample-estimate cycles within the prediction interval, substantially improves system robustness while reducing per-frame processing time and enhancing computational efficiency.

There are still some limitations. First, the paper discusses performance in scenarios involving bumps or rapid motion, and further extension to highly dynamic and occluded environments is needed to achieve a more versatile and robust pose estimation system. Second, although the re-estimation policy significantly enhances system robustness, its effectiveness may be compromised in severely sensor-degraded environments. Addressing these limitations constitutes a direction for future work.

%\begin{IEEEbiography}[{\includegraphics[width=1in,height=1.25in,clip,keepaspectratio]{fig1.png}}]{IEEE Publications Technology Team}
%In this paragraph you can place your educational, professional background and research and other interests.
%\end{IEEEbiography}

\bibliographystyle{ieeetr}
\bibliography{citefile}

\end{document}